\documentclass{article}
\usepackage{graphicx} 
\usepackage{hyperref}
\usepackage[margin=1.25in, top=1in, bottom=1.5in, right=1.5in, left=1.5in, headheight=15pt]{geometry}
\usepackage{booktabs}

\title{FLORIDA: Fake-looking Real Images Dataset}
\author{ali borji\\
aliborji@gmail.com}
\date{October 2023}

\begin{document}

\maketitle

\begin{abstract}
Although extensive research has been carried out to evaluate the effectiveness of AI tools and models in detecting deep fakes, the question remains unanswered regarding whether these models can accurately identify genuine images that appear artificial. In this study, as an initial step towards addressing this issue, we have curated a dataset of 795 genuine images that exhibit a fake appearance and conducted an assessment using two AI models. We show that these models exhibit subpar performance when applied to our dataset. Additionally, our dataset can serve as a valuable tool for assessing the ability of deep learning models to comprehend complex visual stimuli. Indeed, we demonstrate that even cutting-edge segmentation and face detection models struggle with certain difficult stimuli present in our dataset. We anticipate that this research will stimulate further discussions and investigations in this area. Our dataset is accessible at~\href{https://github.com/aliborji/FLORIDA}{https://github.com/aliborji/FLORIDA}.

\end{abstract}

\section{Introduction}

In the realm of visual perception, reality is a complex and often elusive concept. What we see, how we interpret it, and the boundaries between the authentic and the illusory can blur in fascinating ways. This intriguing interplay between the real and the surreal forms the foundation of our exploration into the world of images that challenge our very understanding of what is genuine. From extraordinary landscapes to manipulated photography, optical illusions to unique lighting, this journey takes us through scenarios where real images unexpectedly and inexplicably look fake. This paper delves into the intricacies of this topic by presenting a dataset of real images that exhibit an appearance of falseness.

Deepfakes are deceptive media created using deep learning techniques, like complex neural networks, to manipulate images and voices. They produce highly realistic but fake content, raising ethical concerns due to their potential for misuse, such as spreading false information or impersonating individuals~\cite{fallis2021epistemic,kietzmann2020deepfakes}. Detecting and countering deepfakes is an ongoing challenge, and media literacy is important to reduce their impact. The term ``deepfake" combines ``deep learning" and ``fake" to describe the technology's ability to create convincing fake content.

AI-based tools for detecting deepfakes have improved but are not perfect. Their effectiveness varies depending on the quality and sophistication of the deepfake. While there have been advancements in AI models and techniques for recognizing patterns associated with deepfake generation, the field is continually evolving. Both detection methods and deepfake creation techniques are in constant development.

Despite extensive research focused on evaluating the effectiveness of AI tools and models in detecting deep fakes, the question remains uncertain as to whether these models can also distinguish real images that appear fake but are authentic. In this study, representing our initial exploration in this area, we have gathered a dataset of genuine images that exhibit a fake-like appearance and subsequently tested two models on this dataset. We anticipate that our study will stimulate more discussions and investigations in this particular direction. The dataset is openly accessible via the provided link.

\subsection{Related Research}

In the field of visual perception, various research areas have explored how individuals perceive and interpret images, as well as the factors influencing their understanding of reality. This includes research on visual illusions, image manipulation detection, the psychology of perception, aesthetics in art, neuroscience, and cultural studies. While there may not be specific research dedicated to ``real images looking fake," it falls within the purview of these broader research domains. Researchers continue to investigate the complexities of human perception and the impact of technology and culture on our understanding of reality.

Testing deepfake detectors against real images that look fake is less common because the primary focus of deepfake detection is to identify artificially manipulated content rather than authentic images that naturally appear fake\footnote{The images within our dataset can be regarded as extreme instances ideal for evaluating the effectiveness of fake-versus-real detection methods.}. However, there may be scenarios where detection tools are used to identify real images that have been mistakenly labeled as fake or manipulated due to unusual or surreal characteristics. These situations may include:

\begin{itemize}
    \item {\bf Artistic or Creative Imagery:} Some images created for artistic or creative purposes intentionally have a surreal or ``fake" appearance. Deepfake detectors are generally not designed for such images, but they may occasionally encounter them in testing scenarios.

\item {\bf Unusual Natural Phenomena:} Images of rare or unusual natural phenomena, such as optical illusions or extreme weather events, can sometimes appear fake. Deepfake detectors are not typically used for identifying such images, as they are focused on detecting digital manipulations.

\item {\bf Forced Perspective:} Images that use forced perspective or other optical tricks to create unusual visual effects may not be the intended targets for deepfake detection.

\end{itemize}

In general, deepfake detectors, such as~\cite{chai2020makes}, are primarily designed to identify digital alterations or synthetic content within media, such as manipulated videos, audio recordings, or images. While they can sometimes encounter authentic but unusual images, their primary function is to address digitally altered or generated media.

If there is a specific need to assess the performance of deepfake detectors in distinguishing between authentic, surreal-looking images and digitally manipulated ones, researchers or organizations would need to design tests or datasets to address this unique use case. However, such testing scenarios may be less common compared to the more typical applications of deepfake detection. Such a dataset is currently missing in this field.

Numerous methodologies have been put forth to address the intricate challenge of detecting deep fake content, as exemplified in research like the work by Chai et al.~\cite{chai2020makes}. Furthermore, some investigations, such as~\cite{borji2023qualitative,borji2022generated}, have delved into the qualitative cues that can serve as effective indicators for discerning whether an image has been generated by artificial intelligence. These cues play a crucial role in unraveling the subtleties of deep fake detection.

For a more comprehensive understanding of the field and a thorough examination of the various methods and strategies employed in deep fake detection, interested readers are encouraged to refer to the following reference materials: Lyu et al.'s work~\cite{lyu2020deepfake}, Tolosana et al.'s research~\cite{tolosana2020deepfakes}, Verdoliva et al.'s contributions~\cite{verdoliva2020media}, Nguyen et al.'s exploration~\cite{nguyen2022deep}, and Barad et al.'s insights~\cite{barad2020image}. These resources provide valuable insights and reviews of the evolving landscape of deep fake detection methodologies and can guide further research in this dynamic field.

\begin{table}[t]
\label{sources}
\begin{footnotesize}
\caption{Sources from which we gathered images}
\begin{tabular}{l}
\toprule
\url{https://www.cracked.com/article_21281_the-top-116-images-you-wont-believe-arent-photoshopped.html}\\ 
\url{https://www.boredpanda.com/confusing-perspectives-pictures/?all_submissions=true&media_id=4080680}\\
\url{https://www.awesomeinventions.com/pictures-of-real-things-which-existence-are-hard-to-believe/}\\
\url{https://brightside.me/articles/21-amazing-photos-that-seem-fake-but-are-actually-real-614660/}\\ 
\url{https://stillunfold.com/omg/10-amazing-pictures-that-are-hard-to-believe-but-are-real}\\
\url{https://www.hiptoro.com/p/20-interesting-photos-that-can-be-hard-to-believe/}\\
\url{https://www.theemergingindia.com/photos-that-are-hard-to-believe-are-real/}\\
\url{https://www.buzzfeed.com/crystalro/nature-photos-look-fake-but-are-real}\\
\url{https://www.visualchase.com/en/real-photos-believe-arent-photoshopped?}\\
\url{https://www.nairaland.com/7742018/some-scientific-photos-hard-believe}\\
\url{https://baklol.com/baks/Weird/15-Images-That-Are-Hard-To-Bel-_1965/}\\
\url{https://knovhov.com/most-amazing-unbelievable-people-in-the-world/}\\
\url{https://www.quora.com/What-are-some-real-pictures-that-look-fake}\\
\url{https://diply.com/6430/24-pics-that-seem-impossible-to-believe} \\
\url{https://hotcore.info/babki/cool-forced-perspective-photos.html}\\
\url{https://www.tiktok.com/@boardroom/video/7127753776454176046}\\
\url{https://www.boredpanda.com/real-not-photoshopped-photos/}\\
\url{https://betterbe.co/life/photos-look-fake-but-real/} \\
\bottomrule
\end{tabular}
\end{footnotesize}    
\end{table}


\begin{figure}
  \centering
  \vspace{-50pt}
  \includegraphics[width=.45\linewidth]{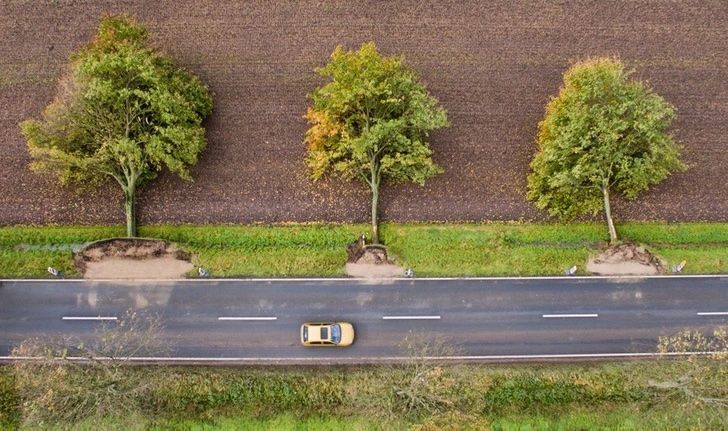}
\includegraphics[width=.45\linewidth]{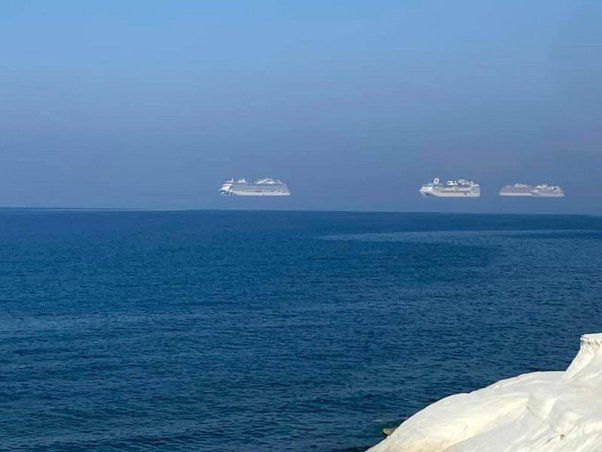}
\includegraphics[width=.45\linewidth]{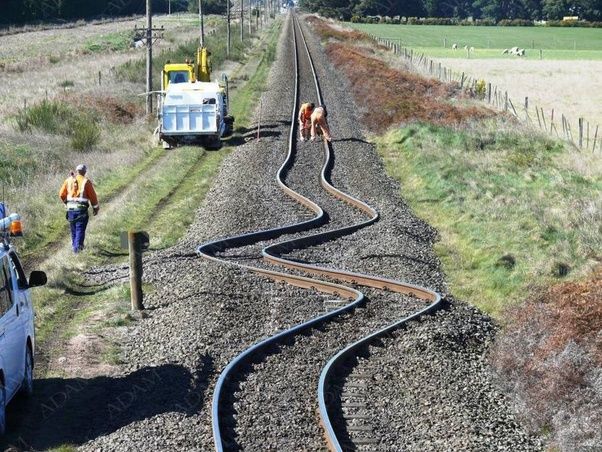}
\includegraphics[width=.45\linewidth]{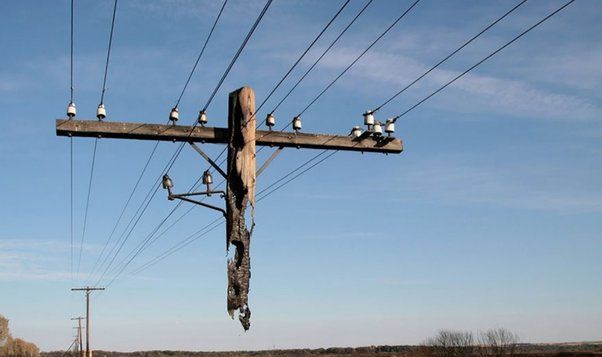}
\includegraphics[width=.45\linewidth]{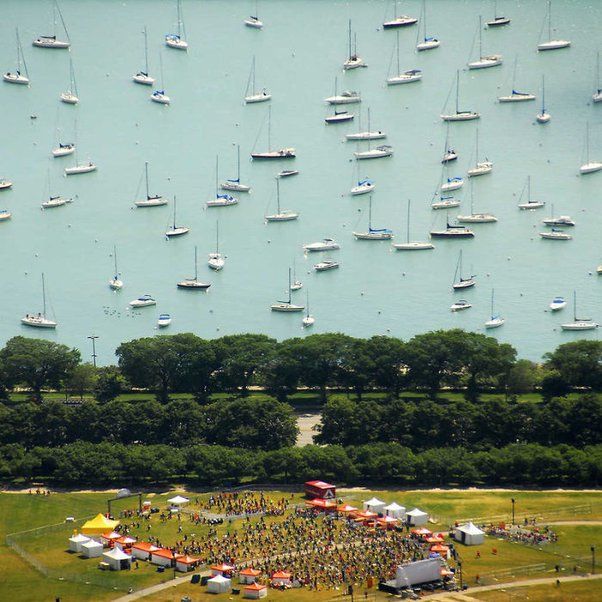}
\includegraphics[width=.45\linewidth]{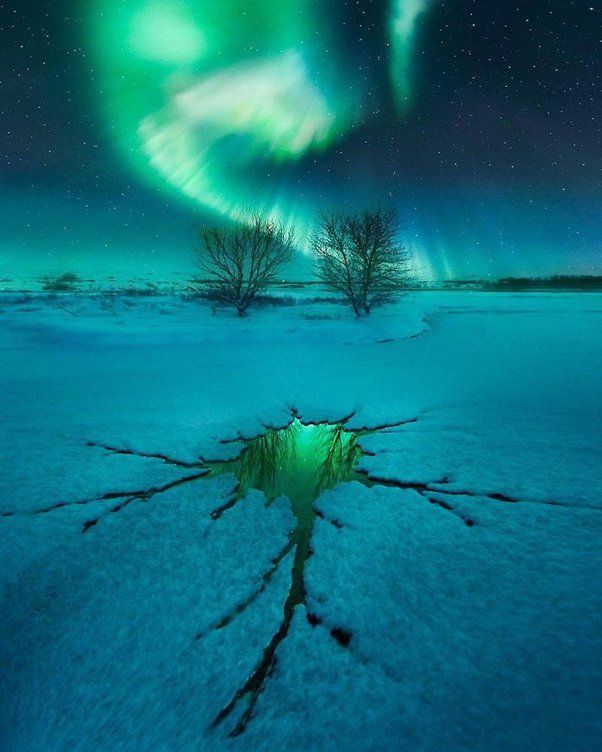}
\includegraphics[width=.45\linewidth]{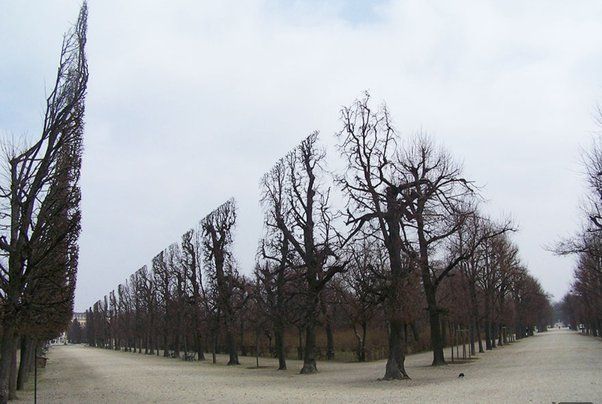}
\includegraphics[width=.45\linewidth]{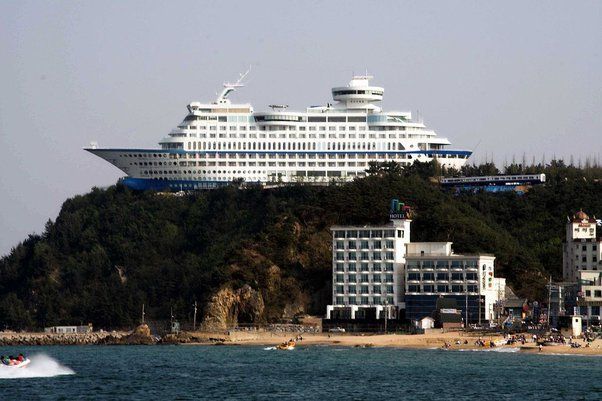}
\includegraphics[width=.45\linewidth]{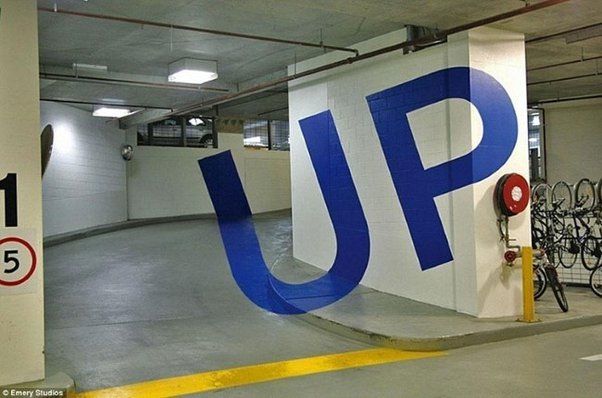}
\includegraphics[width=.45\linewidth]{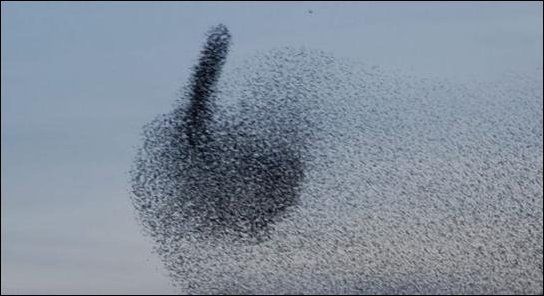}
  \caption{Sample images from our dataset.}
  \label{fig:samples1}
\end{figure}

\begin{figure}
  \centering
  \vspace{-50pt}
  \includegraphics[width=.45\linewidth]{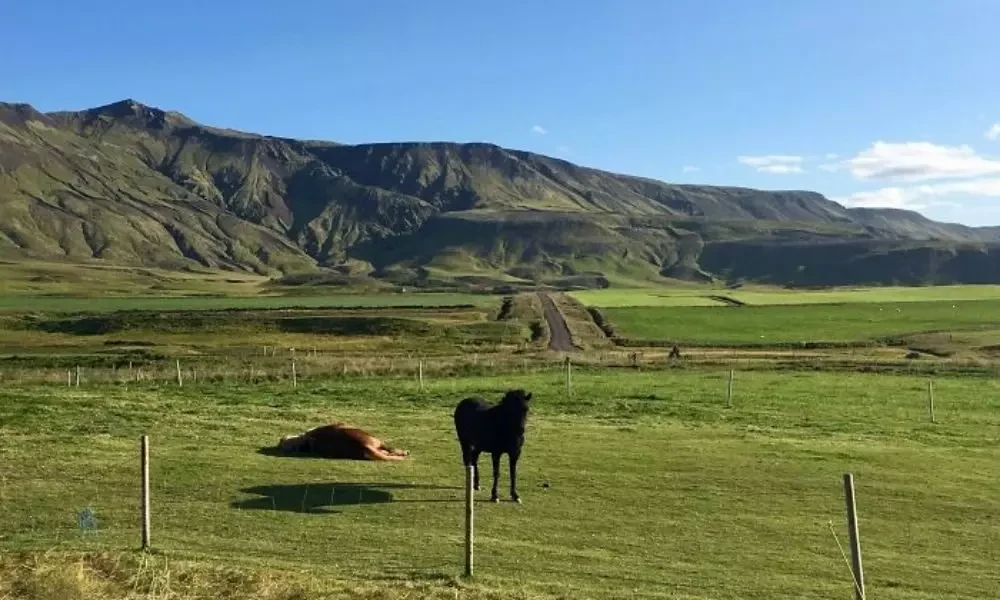}
  \includegraphics[width=.45\linewidth]{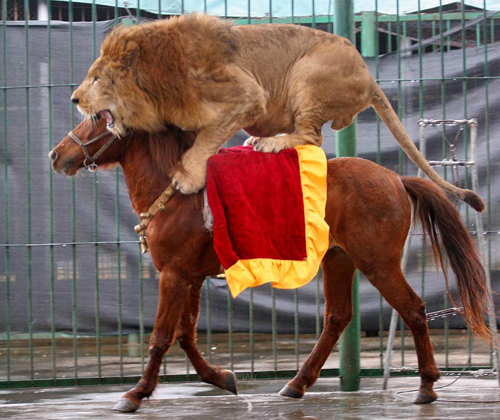}
\includegraphics[width=.45\linewidth]{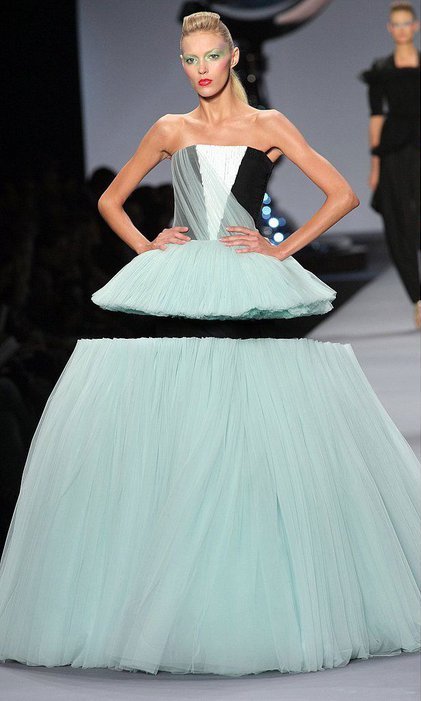}
\includegraphics[width=.45\linewidth]{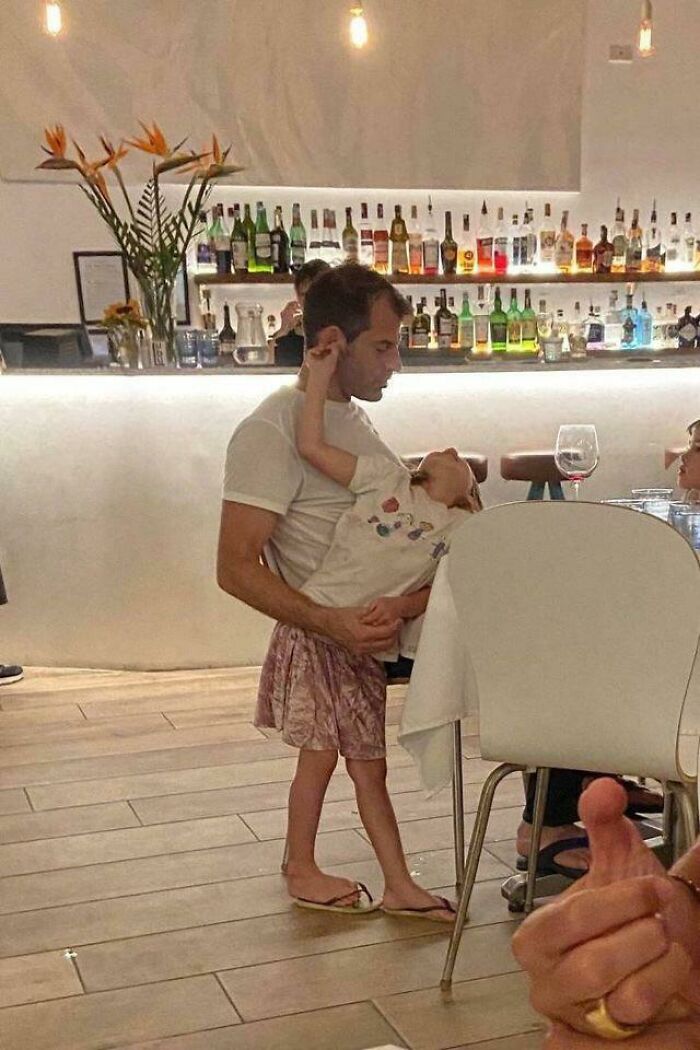}
\includegraphics[width=.45\linewidth]{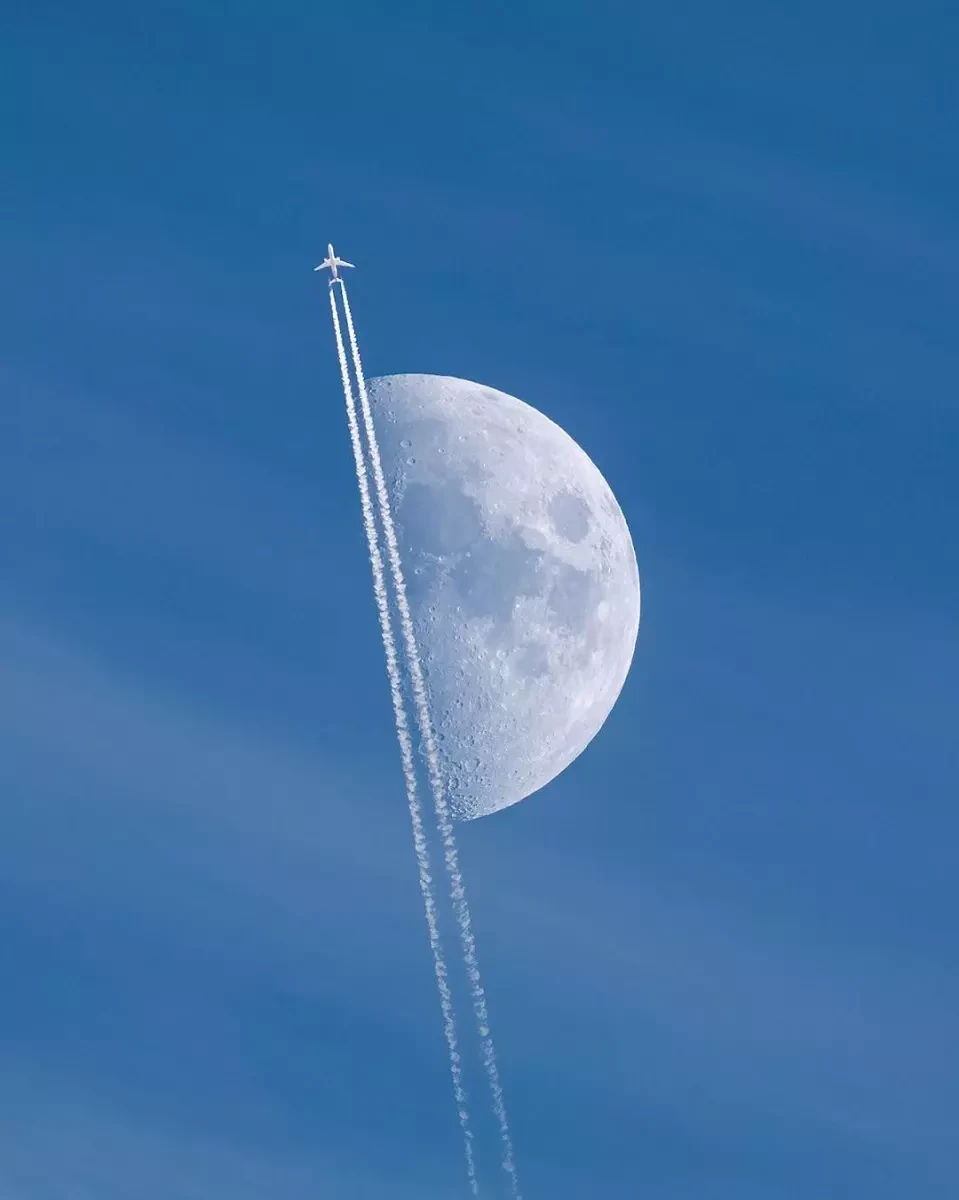}
\includegraphics[width=.45\linewidth]{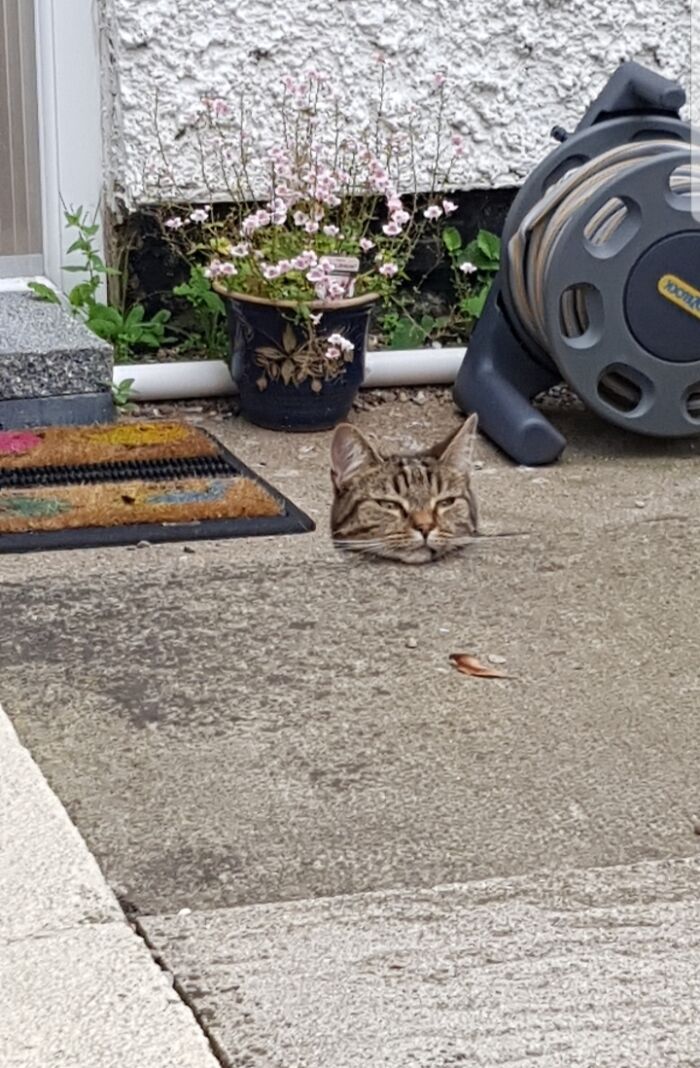}
\caption{Sample images from our dataset (cnt'd).}
  \label{fig:samples2}
\end{figure}

\begin{figure}
  \centering
  \vspace{-10pt}
\includegraphics[width=.45\linewidth]{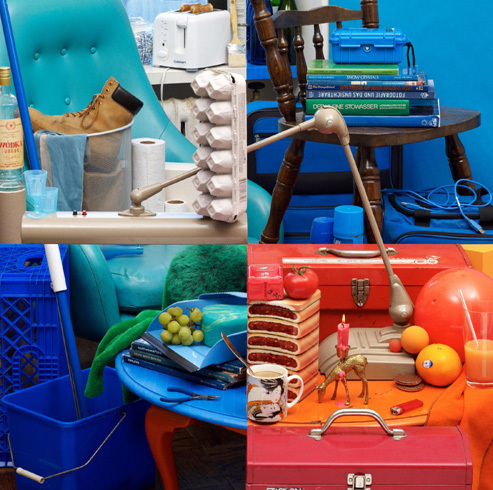}
\includegraphics[width=.45\linewidth]{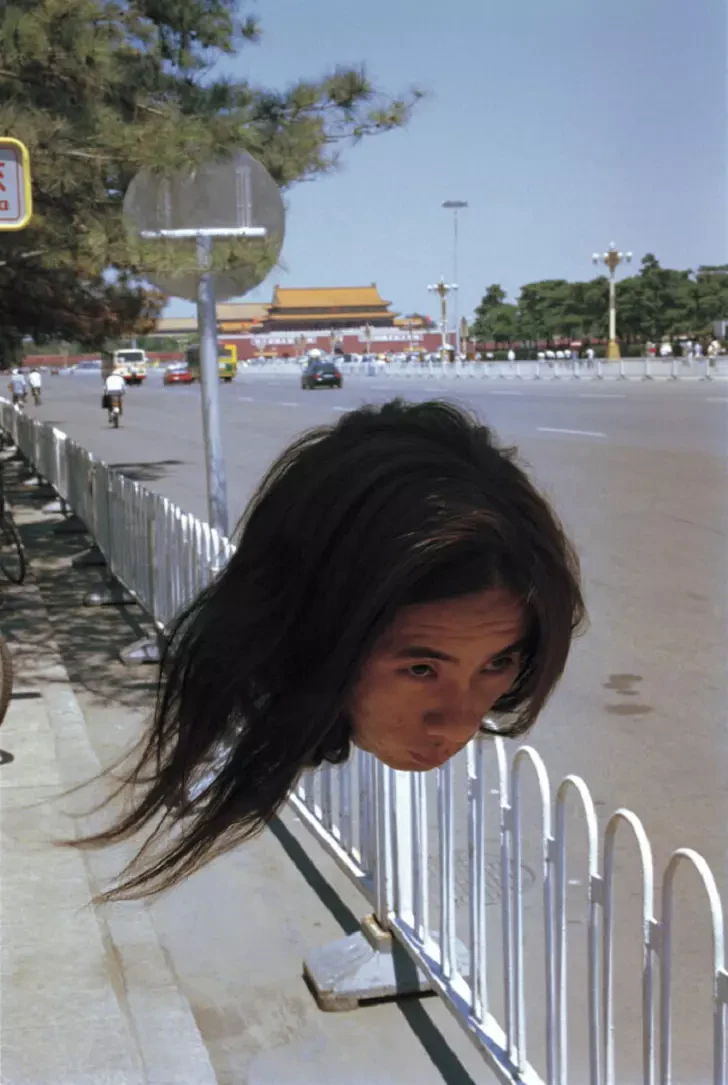}
\includegraphics[width=.45\linewidth]{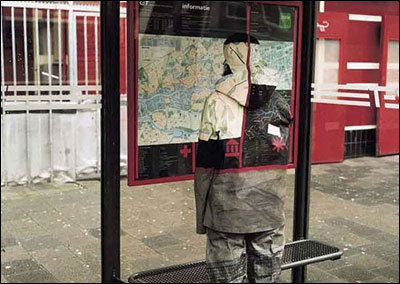}
\includegraphics[width=.45\linewidth]{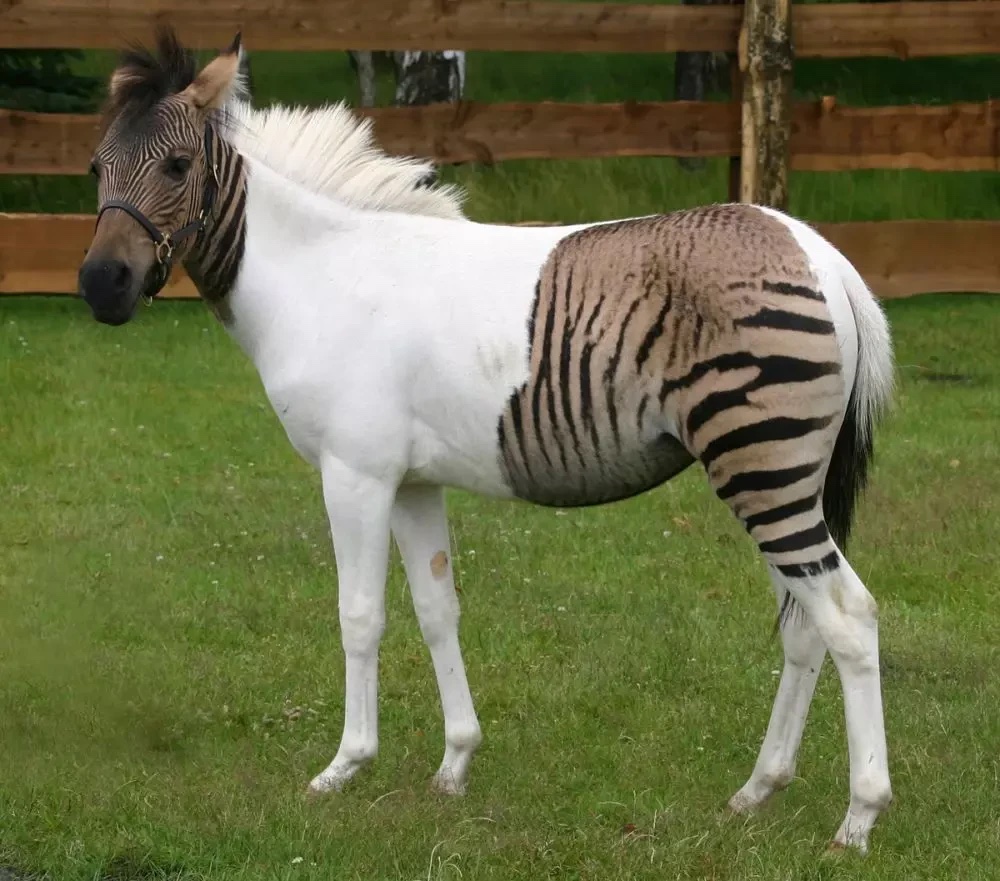}
\includegraphics[width=.45\linewidth]{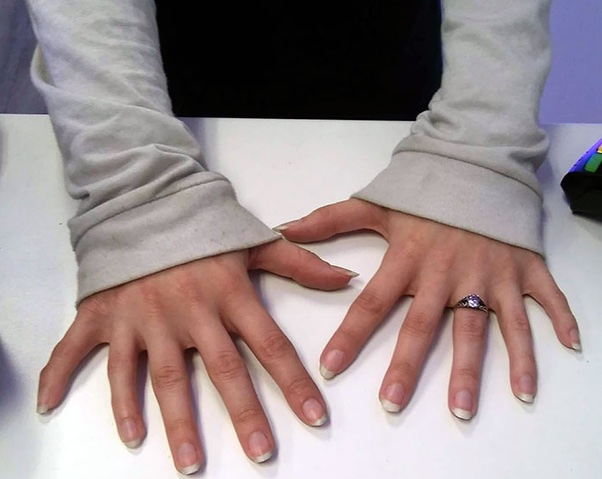}
\includegraphics[width=.45\linewidth]{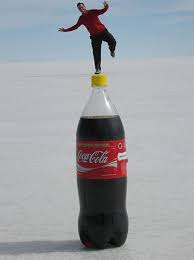}
\caption{Sample images from our dataset (cnt'd). The top right one is taken using a mirror!} 
  \label{fig:samples3}
\end{figure}

\section{FLORIDA Dataset}

\subsection{Fake-looking real images}

Real images can appear completely surreal or fake due to unusual natural phenomena, optical illusions, or creative photography techniques. Examples of real images that might look totally fake are shown in Figures~\ref{fig:samples1},~\ref{fig:samples2}, and~\ref{fig:samples3}. These images may represent instances of the following phenomena:

\begin{itemize}

\item {\bf Mirages:} Atmospheric conditions can create mirages in deserts or over hot surfaces, making distant objects appear distorted or floating in the air.

\item {\bf Fire Rainbows:} Fire rainbows, or circumhorizontal arcs, are optical phenomena that can make clouds look like they're on fire with vibrant, rainbow-like colors.

\item {\bf Light Pillars:} These vertical columns of light occur when ice crystals in the atmosphere reflect and refract light, creating a stunning visual effect.

\item {\bf Bioluminescent Algae:} In certain parts of the world, bioluminescent algae in the ocean can create a surreal, otherworldly glow in the water when disturbed.

\item {\bf Waterspouts:} Waterspouts are tornadoes over water and can appear as twisting, funnel-shaped columns extending from the sky to the water's surface.

\item {\bf Forced Perspective Photography:} Clever use of perspective and camera angles can make real scenes or objects look surreal or out of proportion. This technique is often used in creative photography.

\item {\bf Reflections and Mirrors:} Reflections in glass, water, or mirrors can create visually intriguing effects that might seem unreal.

\item {\bf Unusual Weather Events:} Rare weather phenomena, such as supercell thunderstorms, can lead to eerie and surreal-looking cloud formations and lightning displays.

\item {\bf Natural Geological Formations:} Some landscapes, like the Giant's Causeway in Northern Ireland, have naturally occurring formations that may seem artificial due to their regular, hexagonal patterns.

\item {\bf HDR Photography:} High Dynamic Range (HDR) photography can create hyper-realistic images by combining multiple exposures, which might give an unnatural look to real scenes.

\item {\bf Long Exposure Photography:} Images captured using long exposure times can create surreal effects, such as light streaks, ghostly figures, or blurred motion.

\item {\bf Optical Illusions:} Certain optical illusions can distort real scenes, making them appear fake or altered.

\end{itemize}

These examples demonstrate that the natural world can sometimes produce visuals that are so extraordinary that they appear to be fabricated. It's a testament to the diversity and wonder of our planet's natural processes and the potential for perception to be tricked by optical effects and rare events.

\subsection{Data collection}

We conducted a web search to find genuine images that have an uncanny and almost unbelievable quality to them, making it challenging for people to accept their authenticity. The sources from which the images are gathered can be found in Table~\ref{sources}.

In total, we gathered 795 such images, a selection of which is presented in Figures~\ref{fig:samples1},~\ref{fig:samples2}, and~\ref{fig:samples3}. These images cover a wide range of subjects and content, including those featuring humans and those without.
These images appear unusual, and consequently, their statistical characteristics might be more akin to those of fake images rather than real ones.

\begin{figure}[t]
  \centering
  \includegraphics[width=1\linewidth]{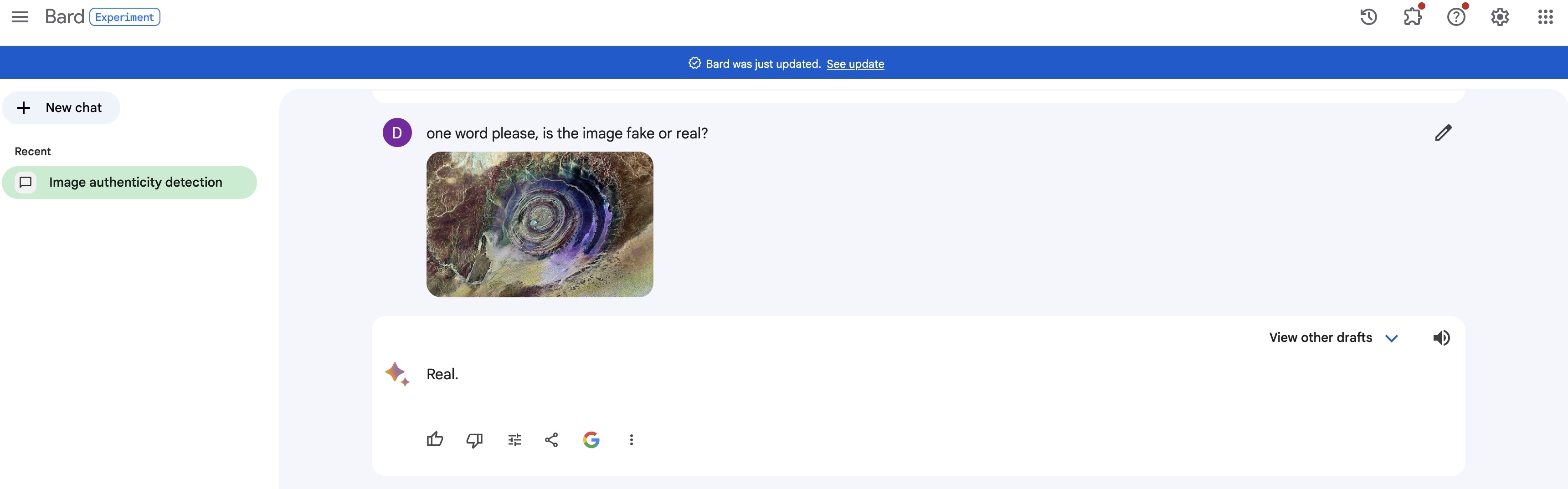}\\
  \vspace{20pt}
  \includegraphics[width=1\linewidth]{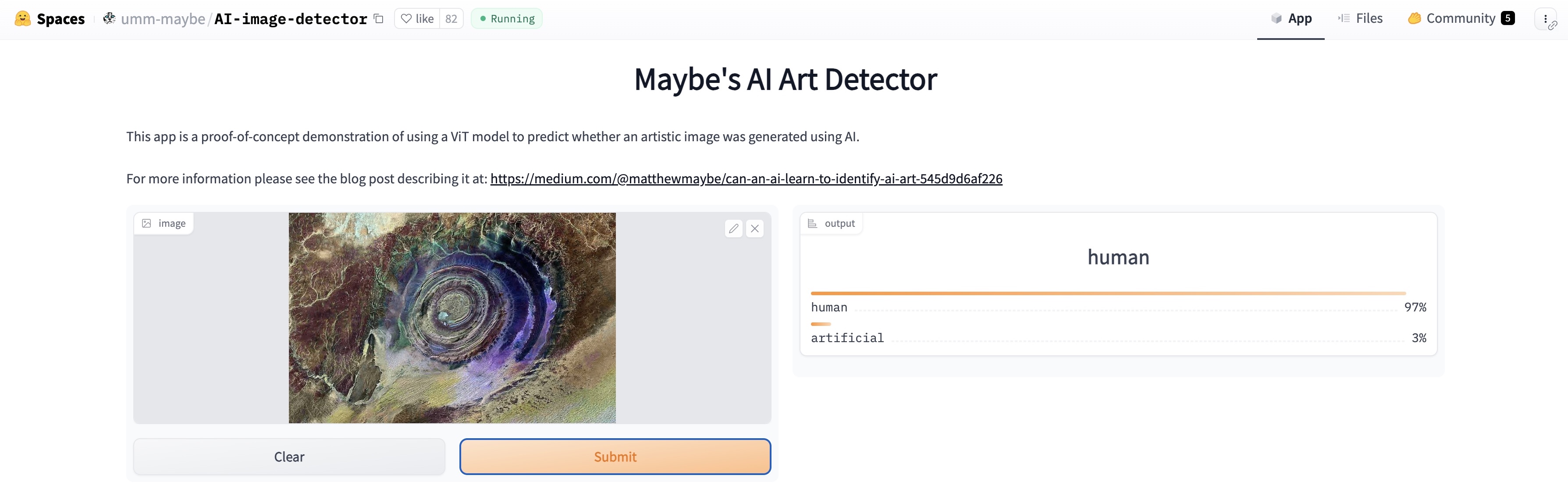}  
  \caption{Online platforms for tested models, top: Bard, bottom: Hugging face API.}
  \label{fig:platforms}
\end{figure}

\begin{figure}[t]
  \centering
  \includegraphics[width=1\linewidth]{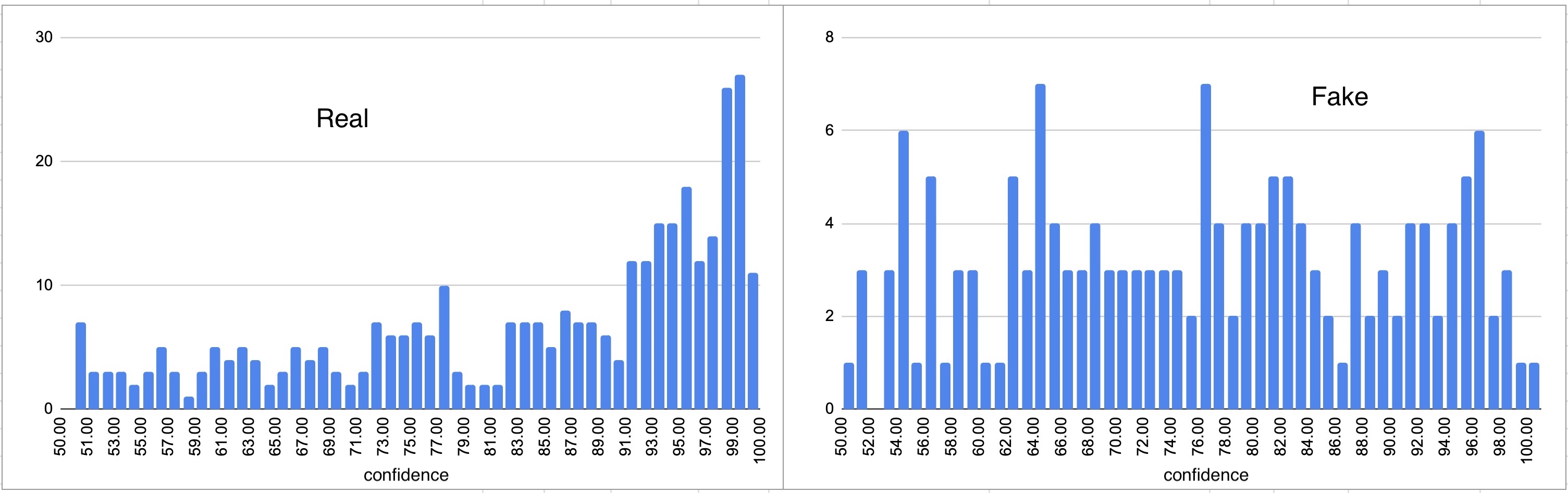}\
  \caption{Confidences of the Hugging face API over FLORIDA images, left: images classified as real, right: images classified as fake.}
  \label{fig:confidence}
\end{figure}

\section{Models and Performance}
\subsection{Real vs. fake detection accuracy}
There exist various models designed for distinguishing between fake and genuine images. In this preliminary investigation serving as a proof of concept, we evaluate two specific models\footnote{This experiment was conducted over 510 images from PART1 and PART2 of the dataset.}. The first model, known as Bard, is a versatile multi-modal large language model (LLM) accessible at \href{https://bard.google.com/}{https://bard.google.com/}. The second model is specialized for detecting fake images and can be found at~\href{https://huggingface.co/umm-maybe/AI-image-detector}{https://huggingface.co/umm-maybe/AI-image-detector}. We manually uploaded the images to these platforms and recorded the results (See Figure~\ref{fig:platforms}).

Bard demonstrates an accuracy rate of 38.2\%, which means it correctly identifies images as real in only 38\% of cases, even though all the images are indeed real. Bard refused to process 92 images that contain people, perhaps due to privacy reasons. In contrast, the Hugging Face API achieves an approximately 67\% accuracy rate. Given that these images look fake for humans the rather high accuracy of the latter model is a bit surprising. The extent of human performance on this dataset remains uncertain and must be evaluated in the future. This could involve, for instance, interspersing these images with other counterfeit ones and soliciting individuals to discern whether an image is genuine or fake.

Confidence distributions of Hugging face API for images classified as real or fake are shown in Figure~\ref{fig:confidence}. It appears that the model tends to provide predictions with greater confidence for real decisions in comparison to fake ones.

There are two notable considerations with respect to these results. Firstly, there is uncertainty about whether these images were part of the training data for the tested models, which could potentially inflate the reported results. Secondly, even in cases where the models performed effectively, they may struggle to provide a clear rationale for their decisions.



\begin{figure}
  \centering
  \vspace{-30pt}
  \includegraphics[width=.24\linewidth]{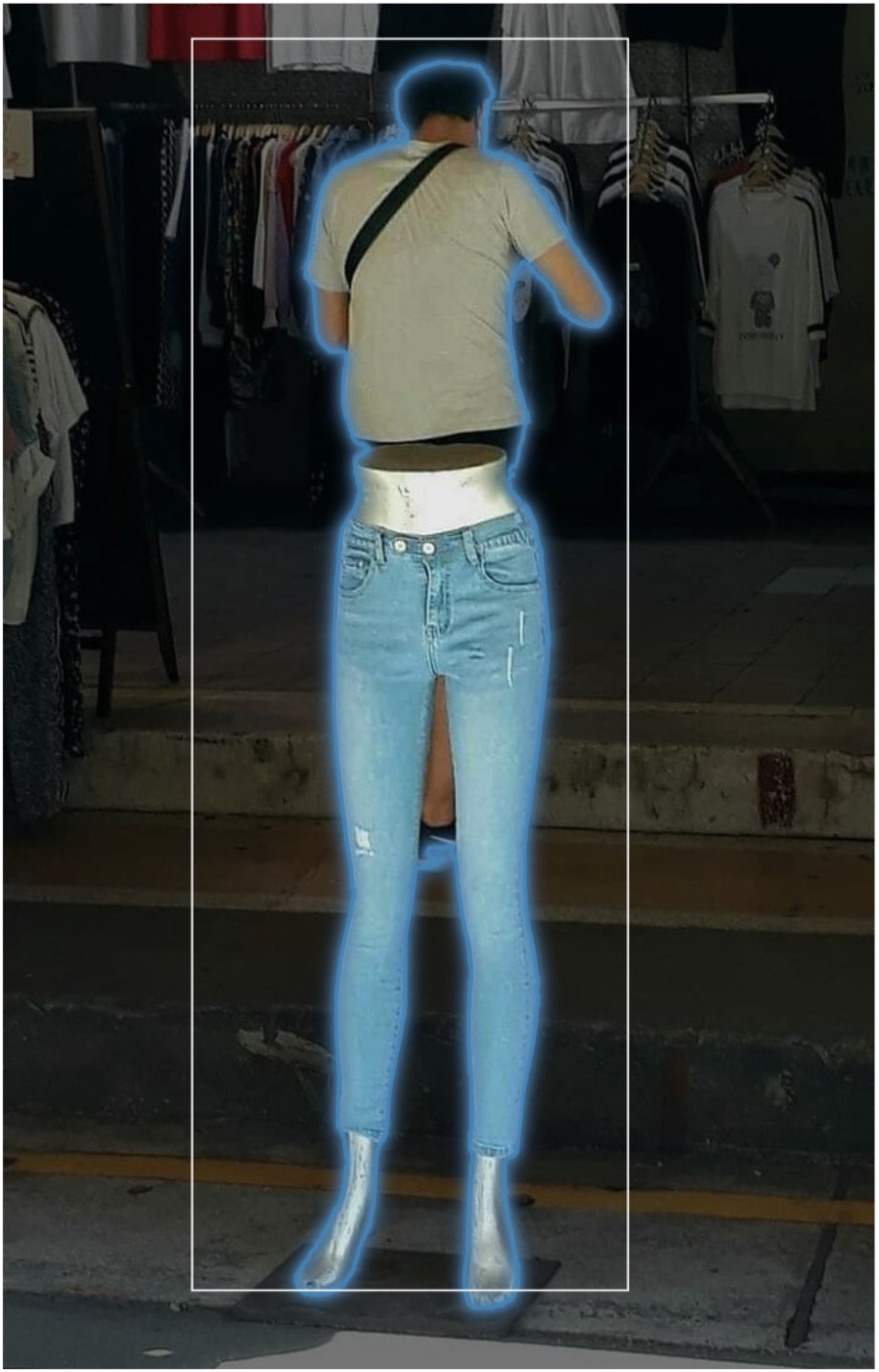}
\includegraphics[width=.24\linewidth]{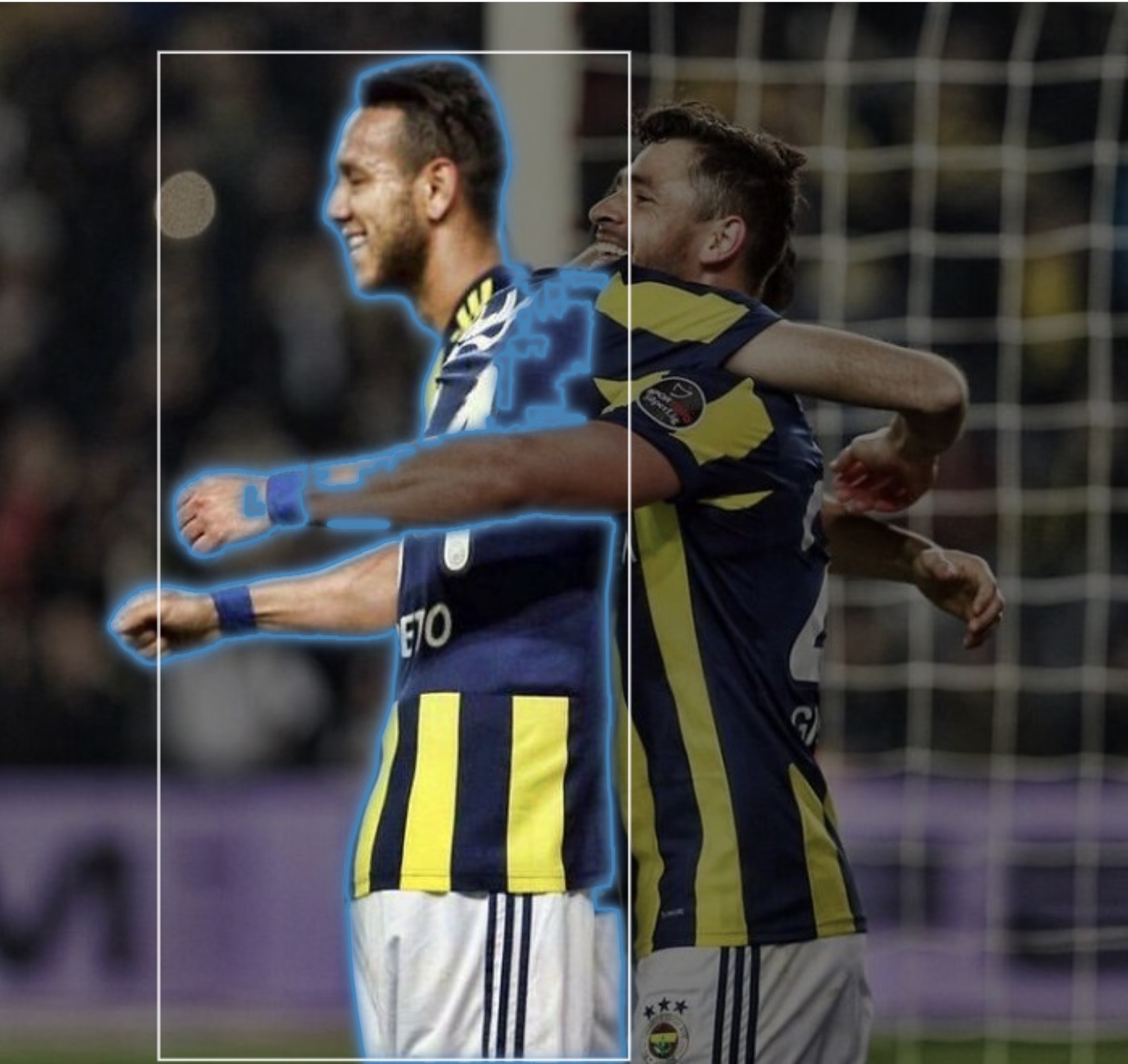}
\includegraphics[width=.24\linewidth]{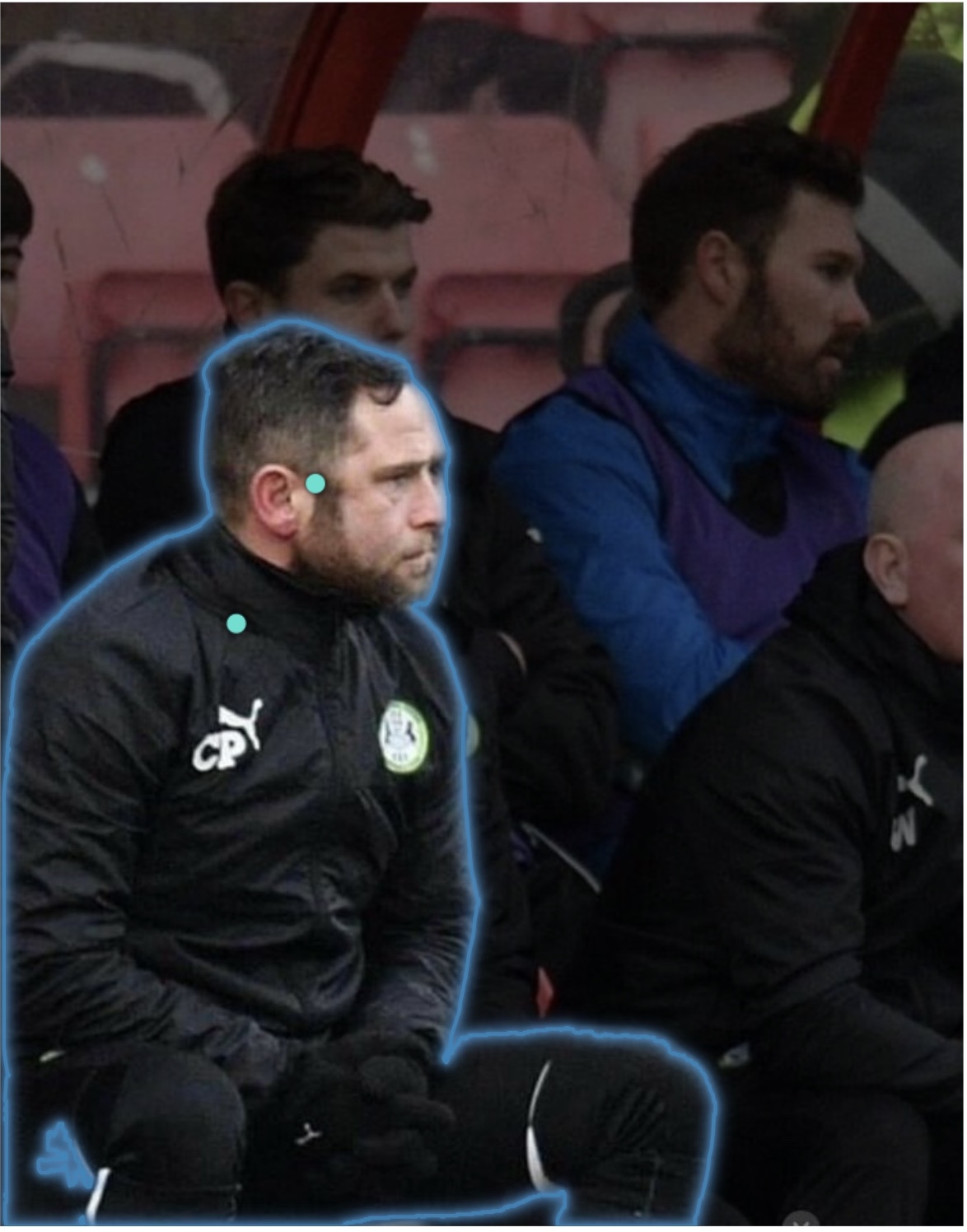}
\includegraphics[width=.24\linewidth]{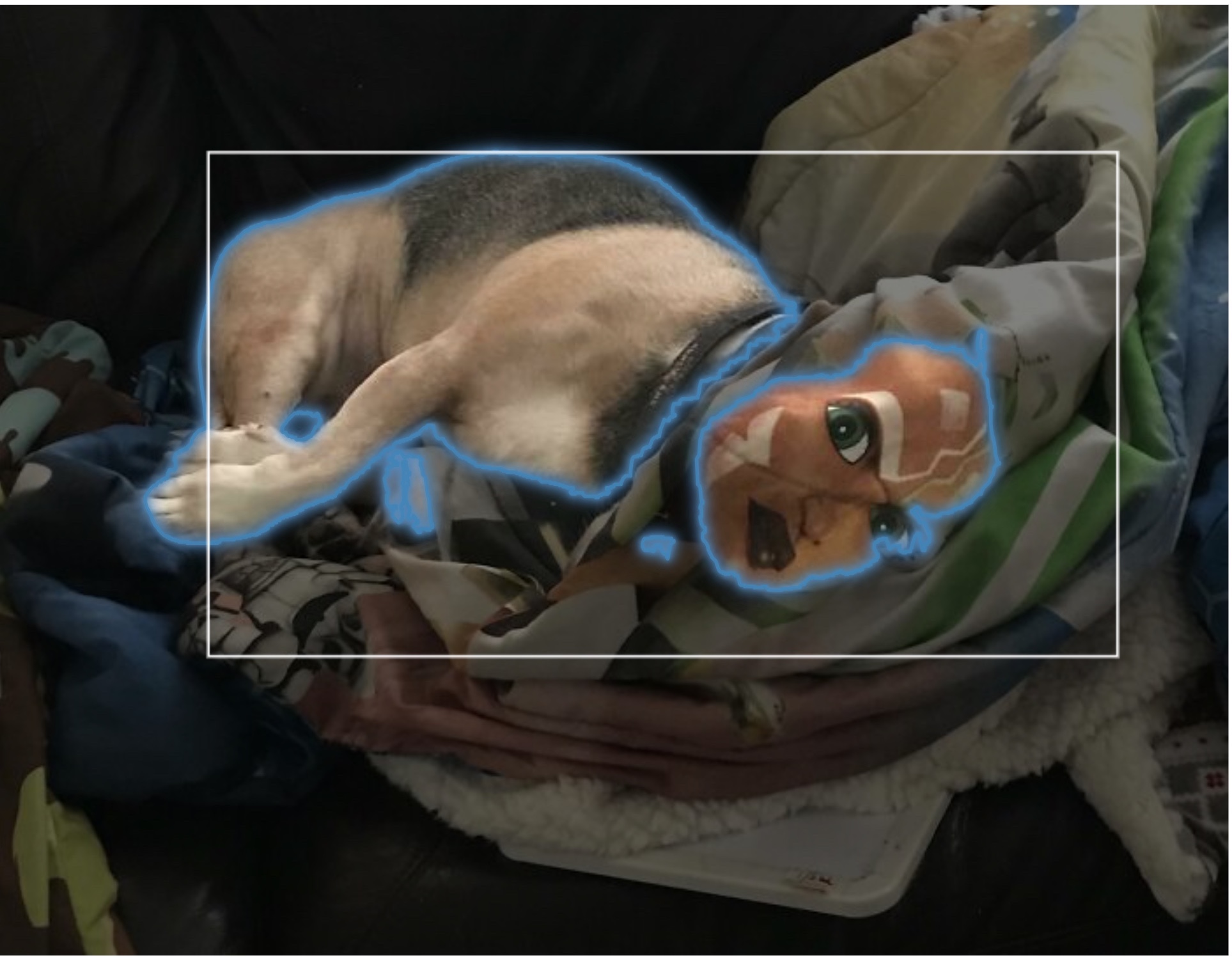}
\includegraphics[width=.24\linewidth]{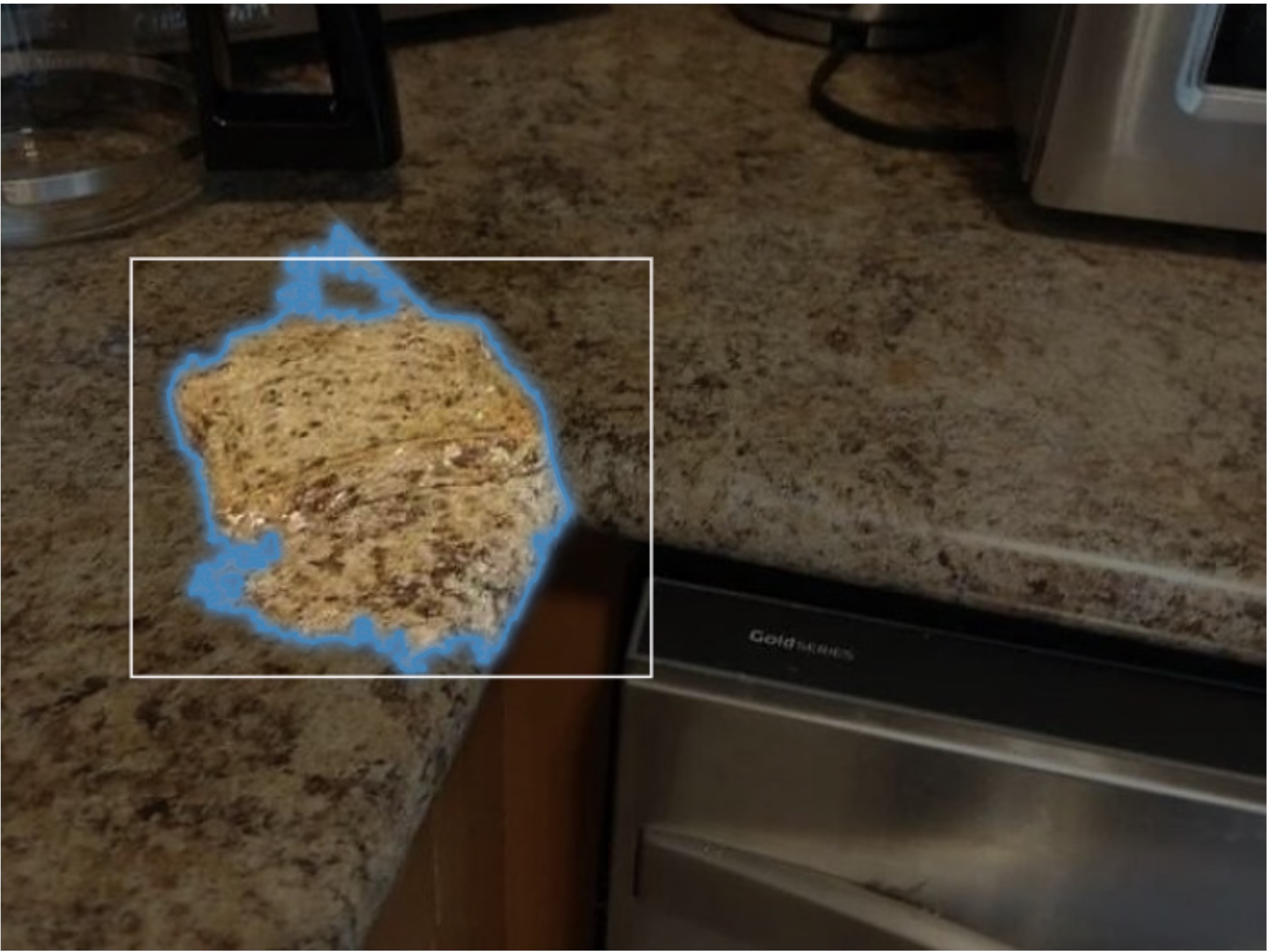}
\includegraphics[width=.24\linewidth]{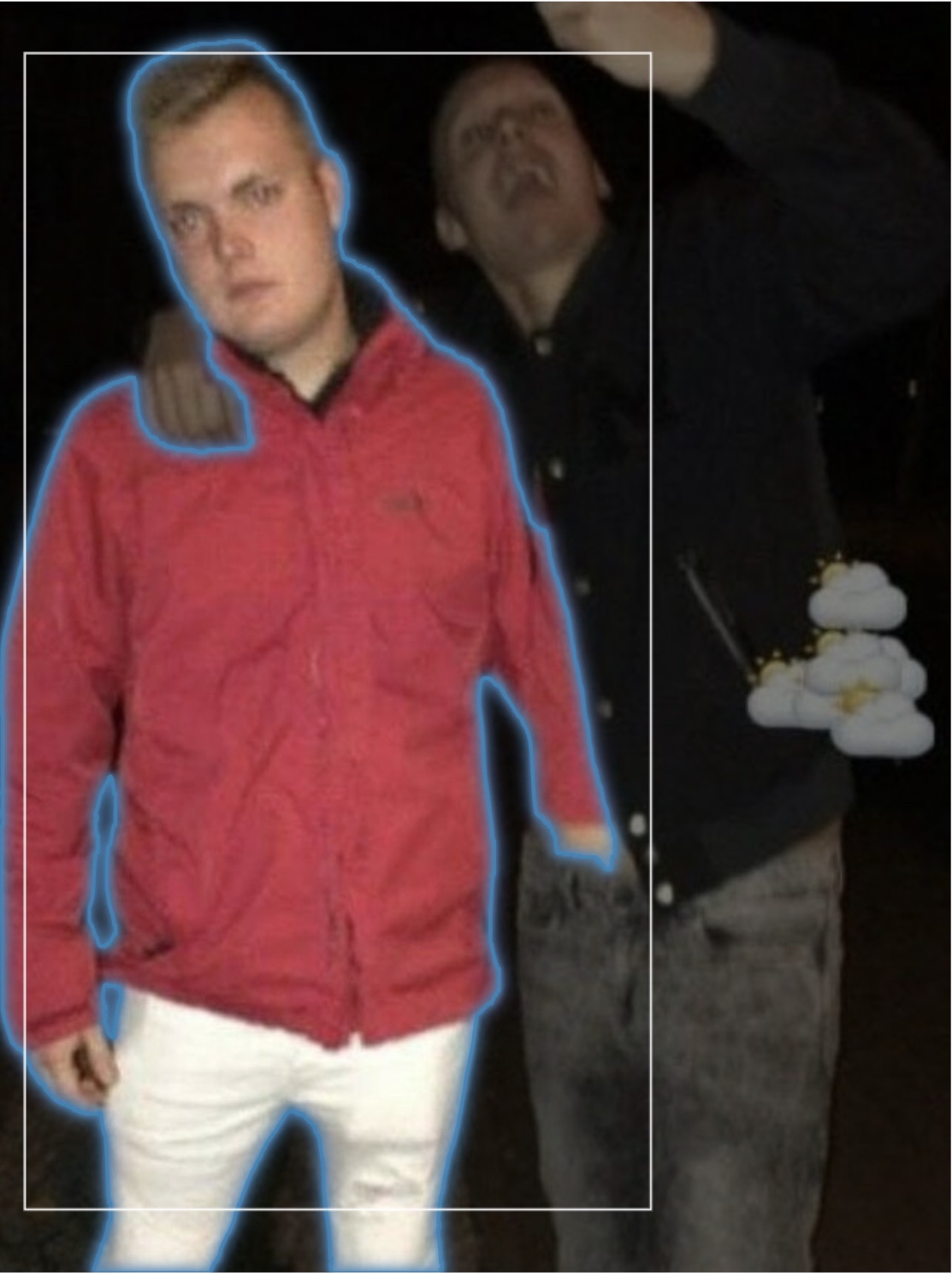}
\includegraphics[width=.24\linewidth]{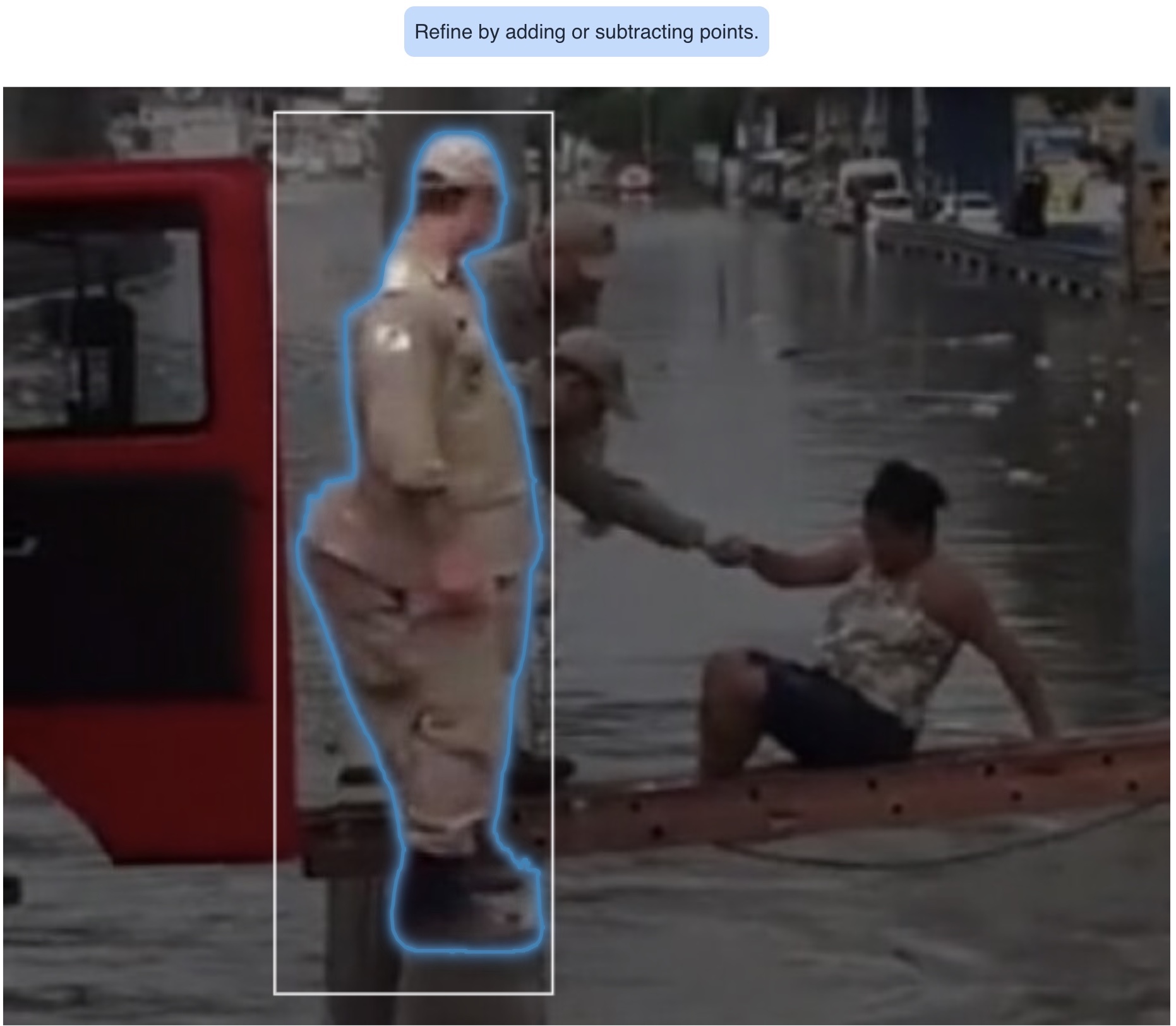}
\includegraphics[width=.24\linewidth]{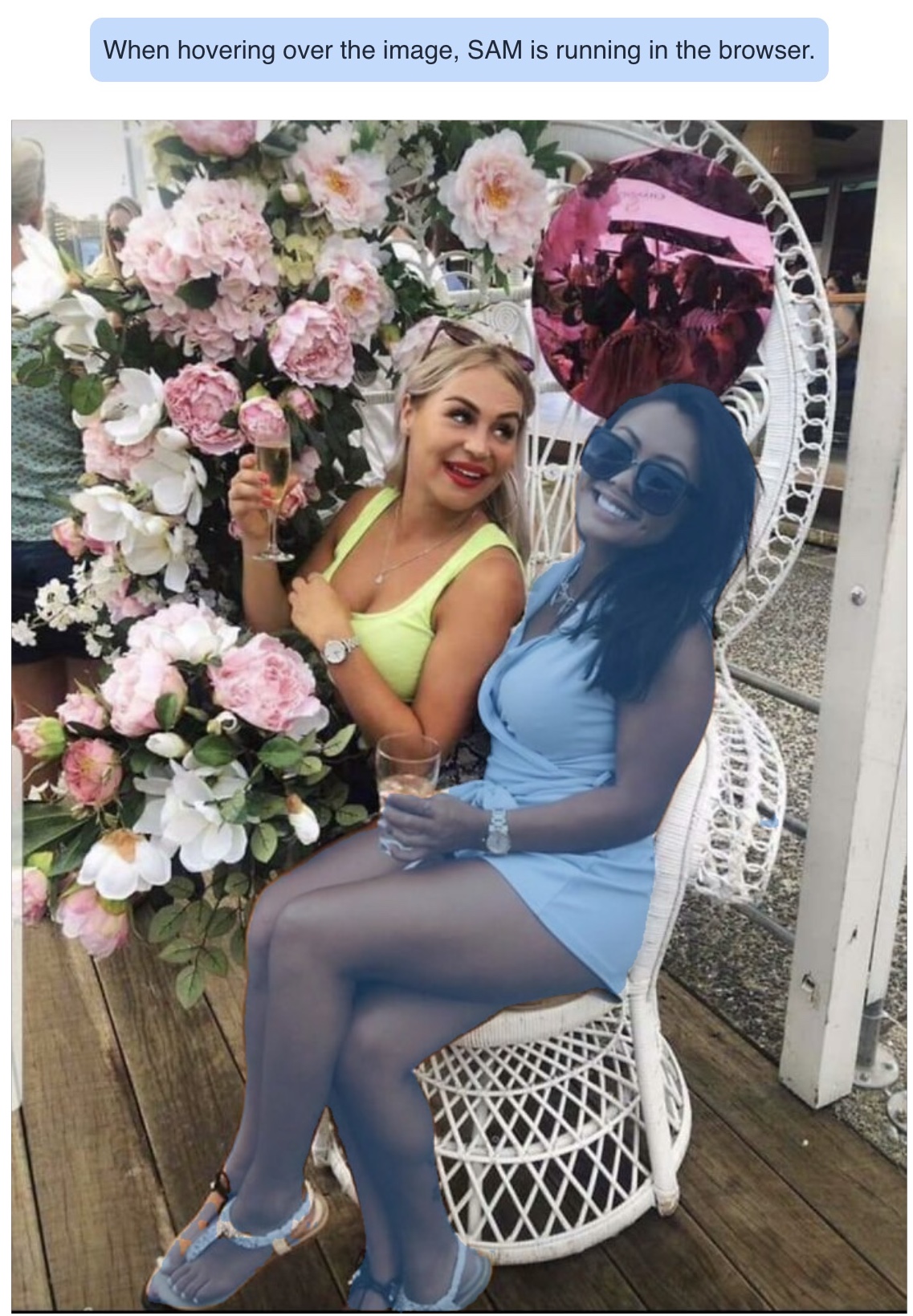}
\includegraphics[width=.24\linewidth]{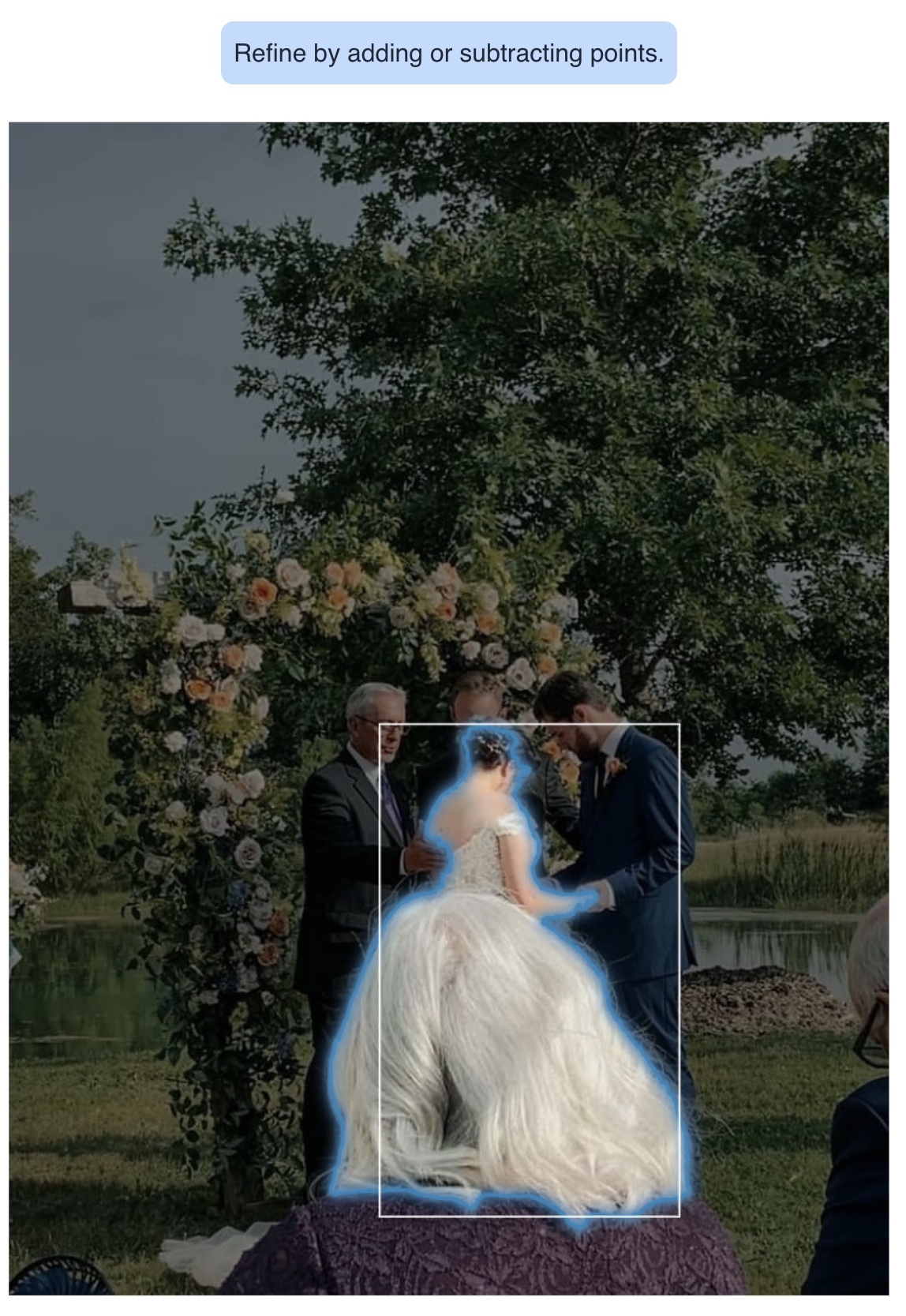}
\includegraphics[width=.24\linewidth]{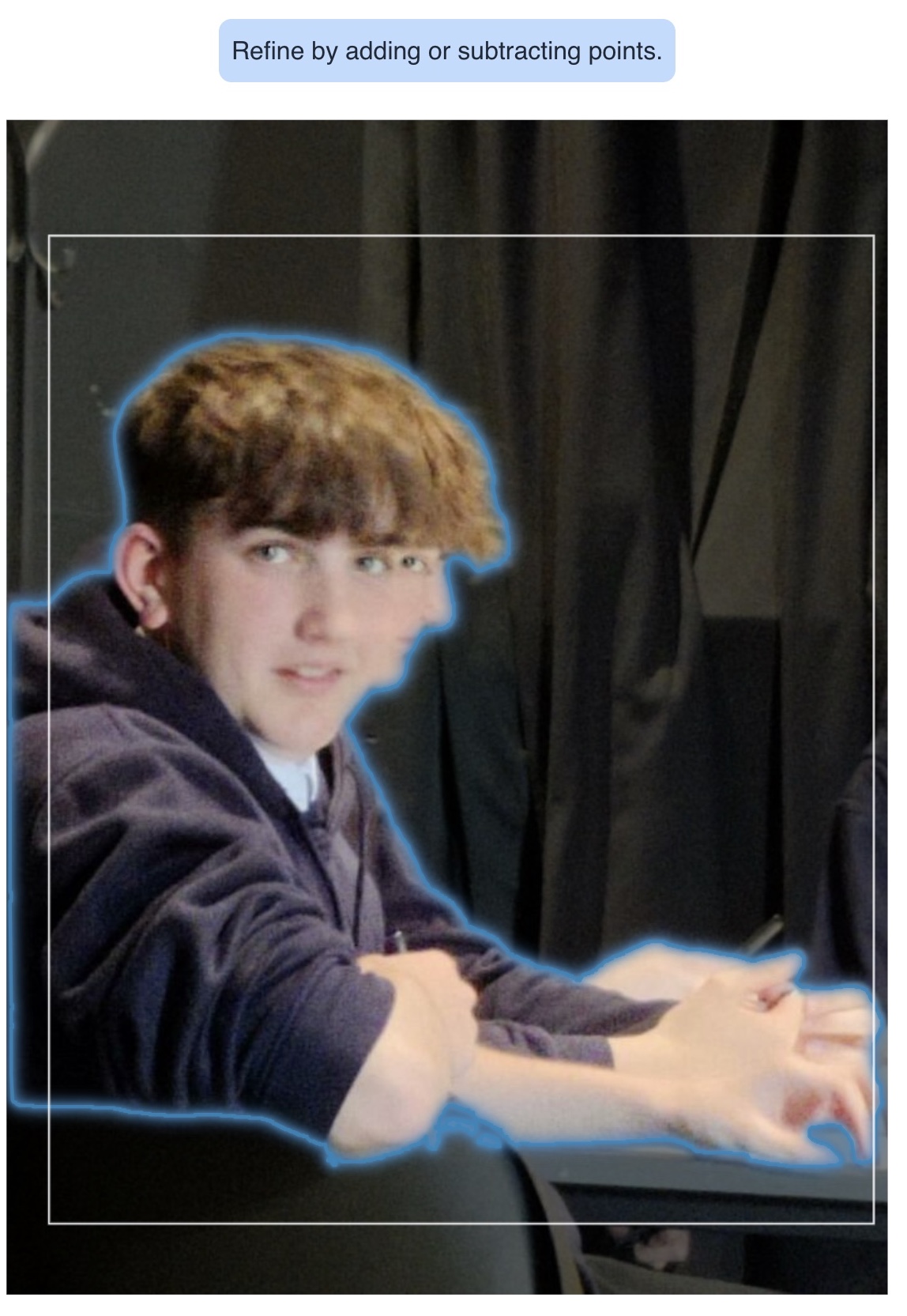}
\includegraphics[width=.24\linewidth]{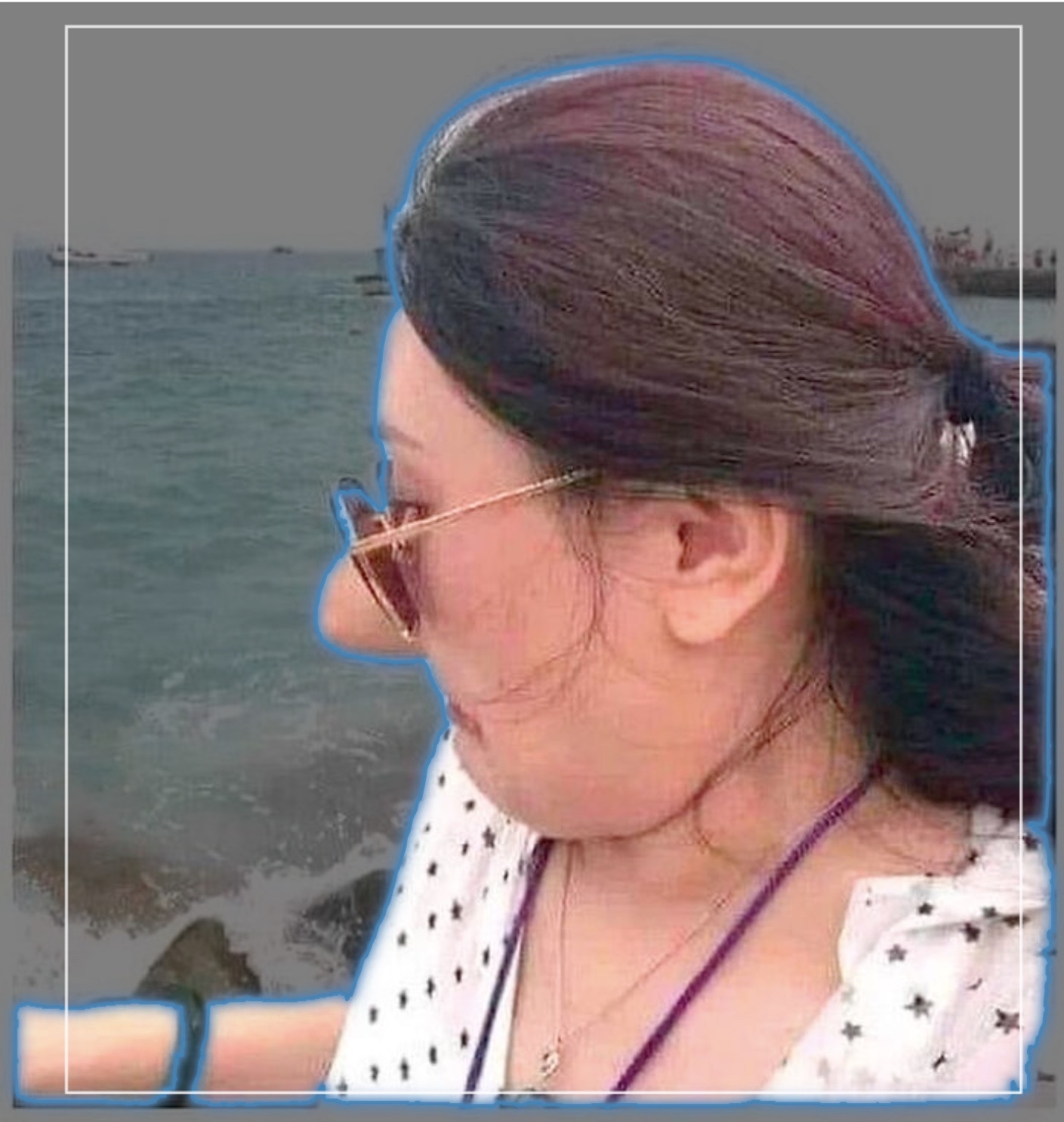}
\includegraphics[width=.24\linewidth]{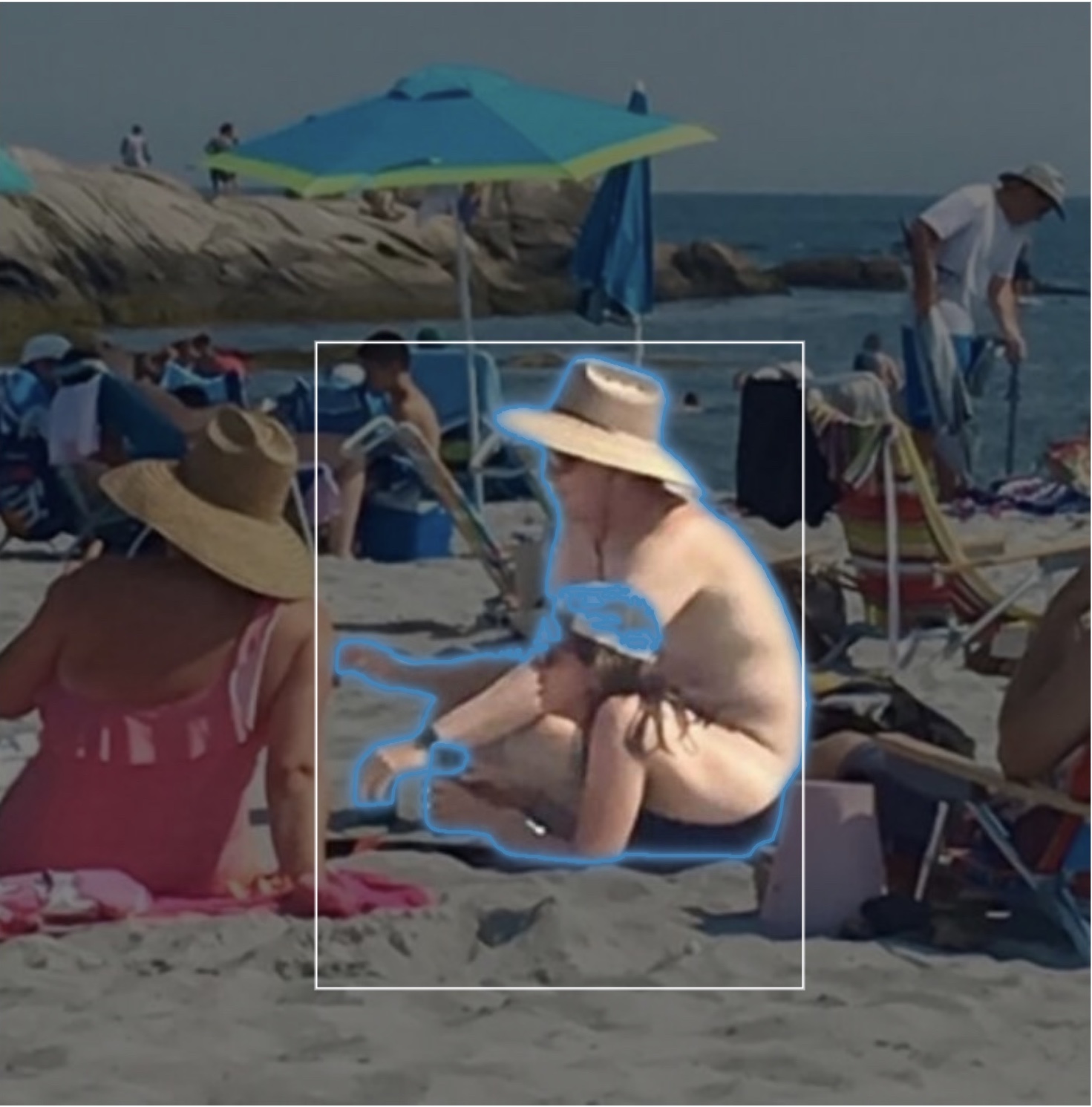}
\includegraphics[width=.24\linewidth]{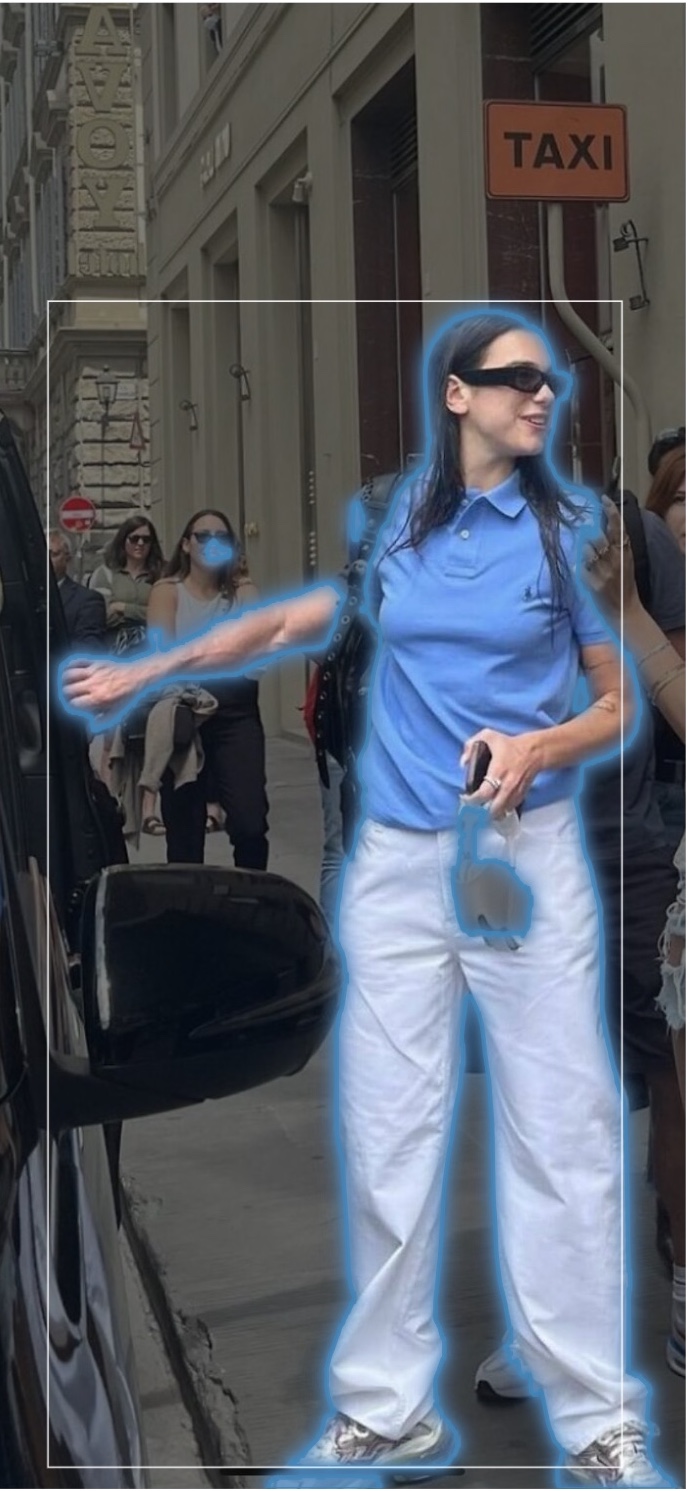}
\includegraphics[width=.24\linewidth]{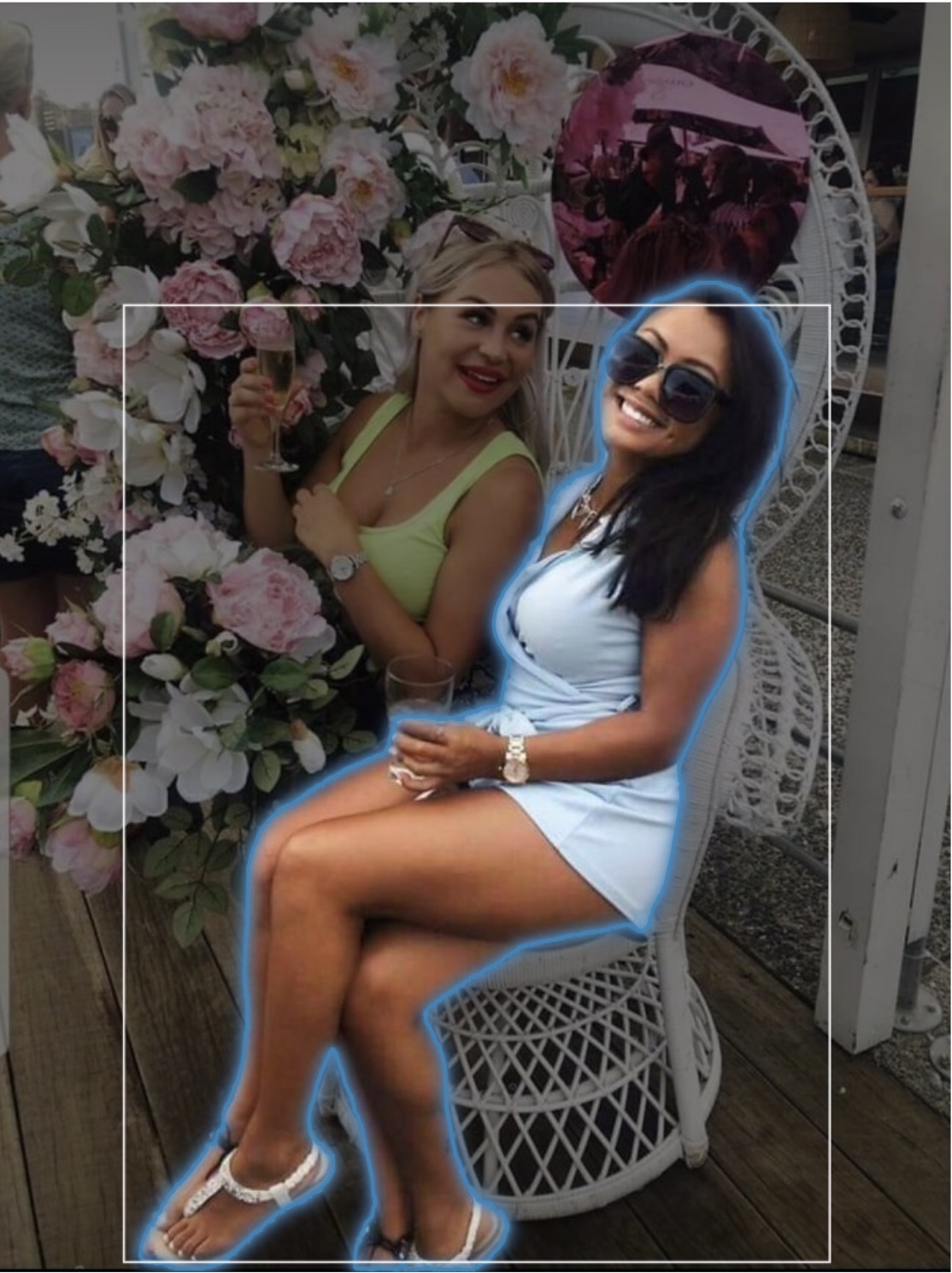}
\includegraphics[width=.24\linewidth]{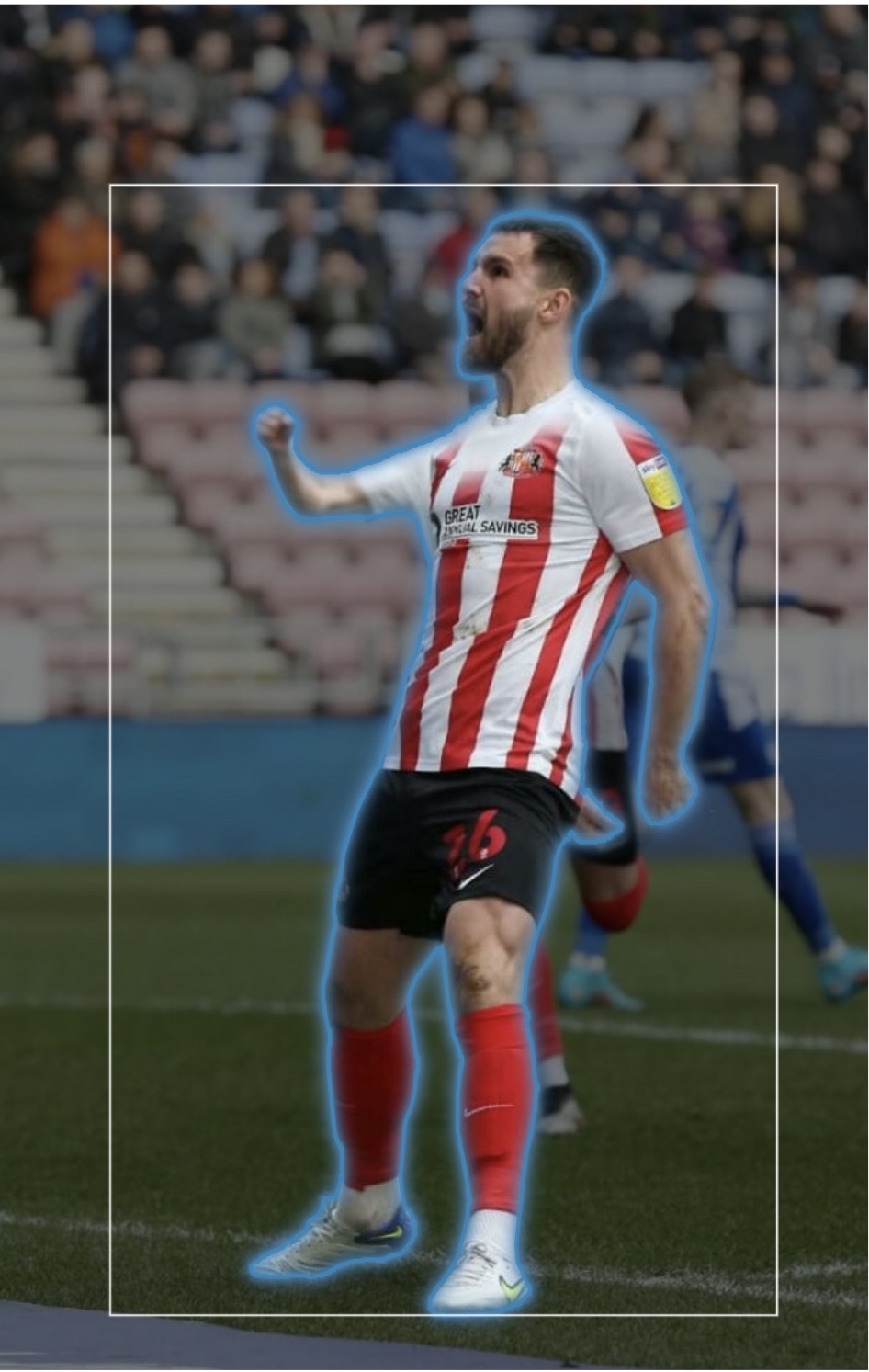}
\includegraphics[width=.24\linewidth]{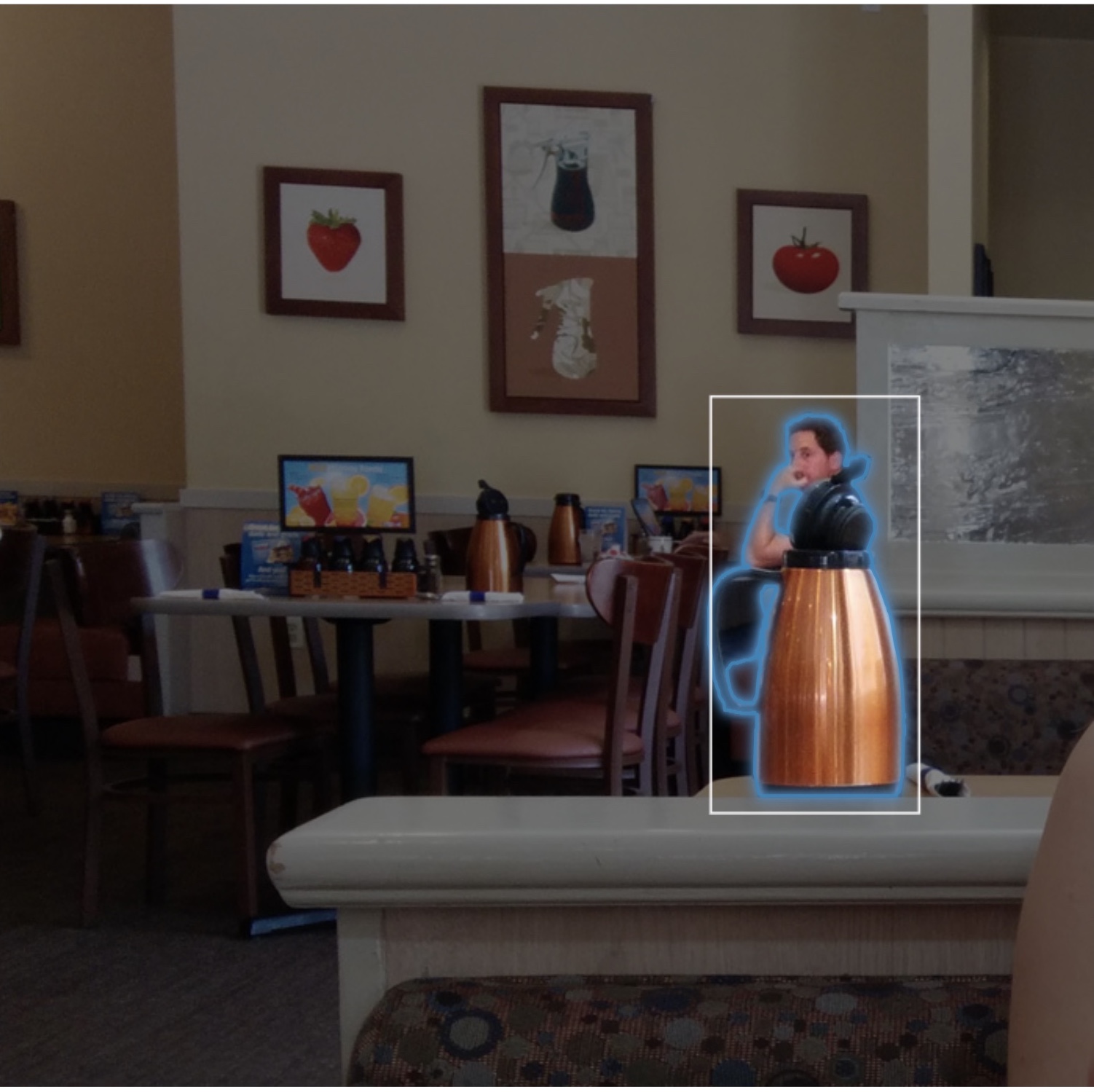}
  \caption{Sample segmentation failure by the SAM model~\cite{kirillov2023segment}.}
  \label{fig:sam}
\end{figure}

\subsection{Performance on other tasks}

Here, we demonstrate the considerable challenge this data poses for models trained in other computer vision tasks, specifically focusing on two tasks: a) semantic segmentation and b) face detection.

We inputted several images into the widely-used SAM (segment anything) model~\cite{kirillov2023segment}, requesting it to delineate the object by outlining a box around the object of interest. The objective was to explore whether this model could accurately isolate objects, such as individuals, which were either overlapping or partially obstructing each other, thereby presenting a scene that is difficult for even humans to interpret. Several instances are illustrated in Figure~\ref{fig:sam}. These findings reveal that this model lacks a profound understanding of real-world physical relationships.

\begin{figure}
  \centering
\includegraphics[width=.32\linewidth]{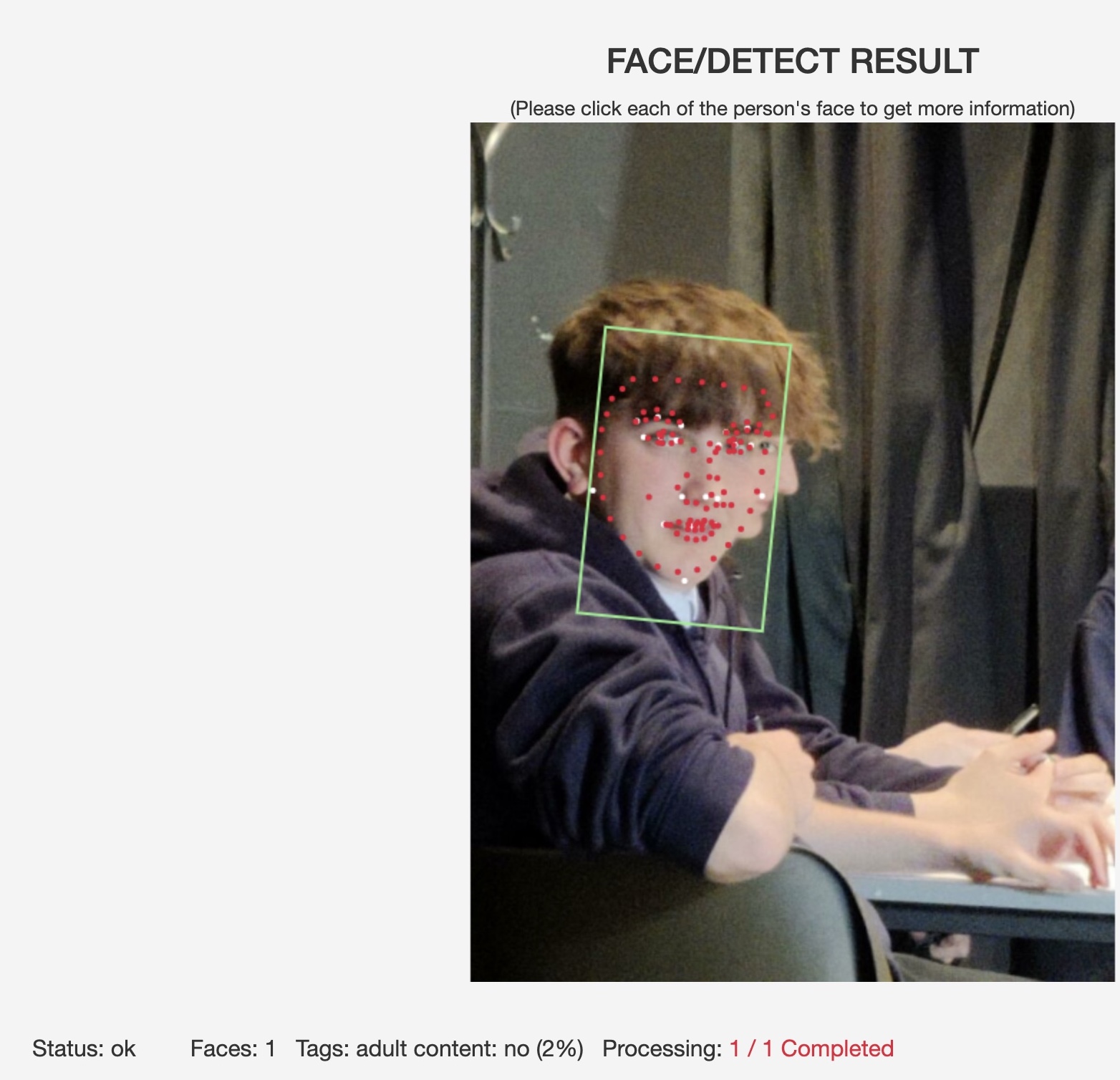}
\includegraphics[width=.32\linewidth]{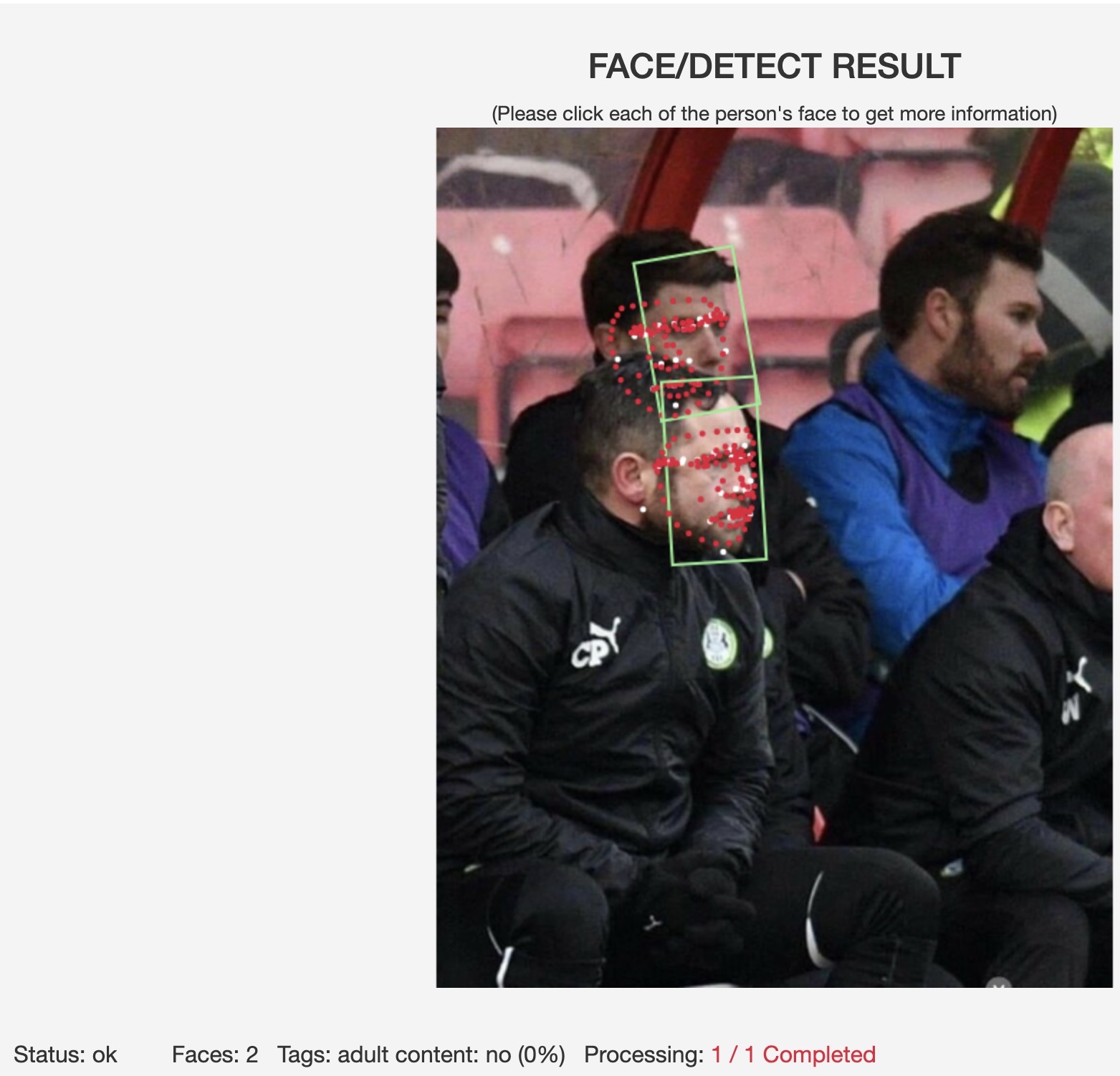}
\includegraphics[width=.32\linewidth]{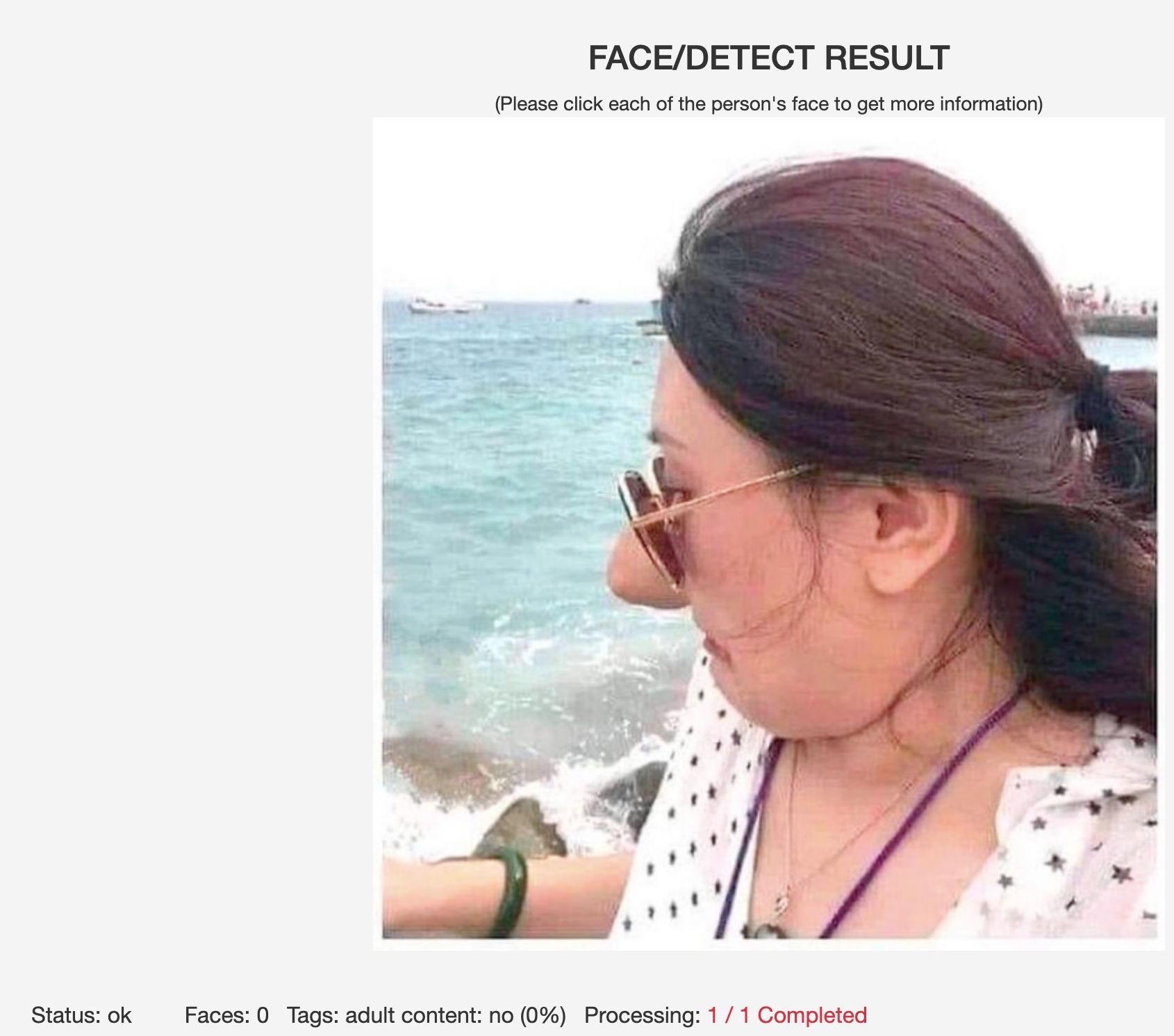}
  \caption{Instances of face detection failures caused by an external tool available at: \url{https://www.betafaceapi.com/demo.html}.}
  \label{fig:faces}
\end{figure}

Additionally, we conducted experiments using an online face detection tool to obtain an initial assessment of its ability to detect overlapping or unconventional faces. As depicted in Figure~\ref{fig:faces}, this tool encountered challenges in detecting faces that were obscured by other faces. In certain cases, it failed to detect the faces within the images.

\section{Discussion and Conclusion}

In our initial exploration of the capabilities of AI deep fake detectors to differentiate between authentic images with a fake appearance and truly fake images, we introduced a dataset and conducted assessments with two models. Our findings indicate that the performance of these models is subpar when confronted with our dataset.

As we move forward, we advocate for the expansion of our dataset and the rigorous evaluation of additional models. We believe that this collaborative effort will foster further advancements in the field of deep fake detection, ultimately leading to more robust and reliable solutions.

The images within our dataset serve as a litmus test for gauging the proficiency of AI models in comprehending the underlying concepts and semantics inherent in images, shedding light on the extent to which these models truly grasp the nuances of the real world. They can be considered as challenging scenarios for state-of-the-art computer vision models and recent multi-modal large language models~\cite{yin2023survey}. They serve as a valuable benchmark to assess the advancements made with the introduction of new models.

An integral aspect and practical application of our dataset lies in its potential to evaluate the depth of comprehension exhibited by AI models when dealing with real-world scenarios. For instance, discerning the authenticity of images in our dataset requires a model (or an observer) to provide a rationale for why an image appears deceptive. Consider, for instance, the second-row image in Figure~\ref{fig:samples2}, where the lady seems to have an incongruity between her upper and lower body. The explanation for this perceptual quirk arises from her attire, which consists of black pants, and her positioning against a black background, resulting in the optical illusion observed.

\bibliographystyle{plain}  
\bibliography{refs.bib}

\end{document}